\documentclass[sigplan,nonacm]{acmart}

\renewcommand\footnotetextcopyrightpermission[1]{}
\usepackage{amsmath}
\usepackage{booktabs}
\usepackage{enumitem}
\usepackage{graphicx}
\usepackage{tabularx}
\usepackage{xspace}

\newcommand{\system}{TomasuLLM\xspace}
\newcommand{\canon}[1]{\operatorname{canon}(#1)}

\newcommand{\inlinesection}[1]{\paragraph{#1}}
\title{TomasuLLM: Out-of-Order Speculative Execution for LLM Agents}
\author{Jiangnan Yu}
\authornote{Jiangnan Yu and Ceyu Xu contributed equally to this work.}
\affiliation{\institution{HKUST}
  \city{Hong Kong}
  \country{Hong Kong}}
\author{Ceyu Xu}
\authornotemark[1]
\authornote{Corresponding author.}
\affiliation{\institution{HKUST}
  \city{Hong Kong}
  \country{Hong Kong}}
\author{Mengming Li}
\affiliation{\institution{HKUST}
  \city{Hong Kong}
  \country{Hong Kong}}
\author{Shiyu Huang}
\affiliation{\institution{Nanjing University}
  \city{Nanjing}
  \country{China}}
\author{Yiran Xia}
\affiliation{\institution{HKUST}
  \city{Hong Kong}
  \country{Hong Kong}}
\author{Jian Weng}
\affiliation{\institution{King Abdullah University of Science and Technology}
  \city{Thuwal}
  \country{Saudi Arabia}}
\author{Hui Xue}
\affiliation{\institution{Zhejiang Lab}
  \city{Hangzhou}
  \country{China}}
\author{Haohui Mai}
\affiliation{\institution{HKUST}
  \city{Hong Kong}
  \country{Hong Kong}}
\author{Zhiyao Xie}
\affiliation{\institution{HKUST}
  \city{Hong Kong}
  \country{Hong Kong}}
\author{Yuan Xie}
\affiliation{\institution{HKUST}
  \city{Hong Kong}
  \country{Hong Kong}}

\begin{document}

\begin{abstract}

Long-running tools can dominate coding-agent latency: compilers, test suites, and repository commands take seconds to minutes while the agent idles. This \textit{observation stall} presents the same tension that drove out-of-order processors---a sequential interface hides work that can be predicted and started early, but a speculative result may become visible only after it and every earlier step have been validated.

We present \system, a runtime that executes agent tool calls out of trajectory order while preserving task-execution correctness. It drafts future actions, runs them in isolated copy-on-write sandboxes, traces their dependencies and effects, and commits results in trajectory order only after validation against committed state.
Across three benchmarks spanning sub-second to minutes-long tool calls, \system improves the reported benchmark means and scales with tool latency: 1.31$\times$ on 100 SWE-bench Verified tasks, 1.35$\times$ on 28 Terminal-Bench 2.0 tasks, and 1.27$\times$ matched progress on 18 SWE-Marathon sessions. Across 4{,}010 audited commit-validation records, it produces zero false accepts.

\end{abstract}

\maketitle

\section{Introduction}
\label{sec:introduction}

Frontier LLMs are becoming execution controllers, not only text generators. Agent harnesses give these models tools, parse the actions they emit, execute those actions, and return observations to the model context~\cite{yao2023react,schick2023toolformer}. In coding agents, the slow part of that loop is often not decoding: model inference now runs at thousands of tokens per second on specialized hardware~\cite{cerebras2026inference,abts2022groq}, while builds, tests, service restarts, and repository commands take seconds to minutes. The result is a runtime bottleneck familiar to systems builders: a fast decision engine repeatedly waits for long-latency operations before it can issue the next useful action.

\begin{figure}[!t]
  \centering
  \includegraphics[width=\columnwidth]{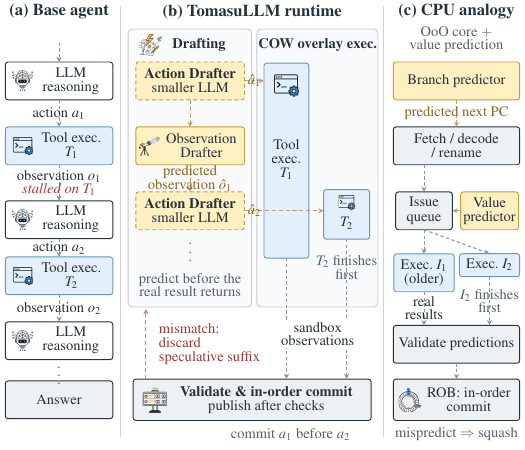}
  \caption{Observation-stall slack and the commit boundary. (a)~The base agent serializes at observation boundaries. (b)~\system executes drafted actions in COW overlays and publishes only validated observations in main-agent order. (c)~The CPU analogy separates prediction from in-order commit.}
  \Description{Three side-by-side panels. Panel a shows the base-agent serial path: LLM reasoning, tool execution T1, LLM reasoning, tool execution T2, and the final answer, where each tool observation must return before the next action issues. Panel b shows the TomasuLLM runtime with a drafting region and an overlay-execution region. The Action Drafter emits predicted tool actions, the Observation Drafter emits a predicted observation, and real tool executions T1 and T2 run in copy-on-write overlays, with T2 finishing first; sandbox observations flow into the Validate and in-order commit stage before publication, and a mismatch discards the speculative suffix. Panel c shows the CPU analogy: branch and value prediction feed fetch, decode, rename, and issue; instructions I1 and I2 execute out of order; validation and the reorder buffer commit in order, with mispredictions squashed.}
  \label{fig:intro-overview}
\end{figure}

Current harnesses serialize around those observations. The main agent emits one or more tool actions, the harness waits for their observations, and only then can the next model response produce later-turn actions. This creates an \textit{observation stall}: the interval in which useful future work may be predictable but cannot be issued by ordinary multi-call dispatch because the action has not yet been decoded. On the 212 Claude/Codex serial reference trajectories published by SWE-Marathon~\cite{swemarathon2026}, the average trajectory has 242 turns and 266 actions, spends 9,315 seconds inside tools, and recovers only 54 seconds from multi-call parallel dispatch. Section~\ref{sec:motivation} shows that the missing slack lies across turns, where same-turn tool parallelism cannot reach it.

Figure~\ref{fig:intro-overview} makes the latency opportunity and correctness boundary concrete. In the base-agent path~(a), each observation must return before the next action can even be decoded, so the wait on $T_1$ hides no younger work. \system breaks that timing dependence~(b) by separating early issue from visible commit. An \textit{Action Drafter} proposes the next tool action from committed history; an \textit{Observation Drafter} predicts what that action will return so the next draft can begin without waiting; and each drafted action executes inside a private copy-on-write (COW) overlay. The main agent remains the authority for canonical actions: a drafted action's observation is returned only after commit validation accepts evidence from the real sandbox execution. Each execution leaves a \textit{Trace IR} record containing the canonical action, the file-version records it read, the observation it produced, and the effects it would publish. Prediction proposes work, the main agent authorizes actions, and commit validation authorizes publication. We call the resulting guarantee \textit{task-execution correctness}: \system may change start times, but it must not publish an observation, artifact, workspace update, or external effect unless that publication is justified by a real execution in the current committed task state.

Prediction makes cross-turn slack reachable, but it does not make reuse correct. The Action Drafter can be a small model fine-tuned as in speculative decoding~\cite{leviathan2023fast,chen2023speculative}, the Observation Drafter a language world model such as Qwen-AgentWorld~\cite{zuo2026qwenagentworld}. Yet a predicted action that looks right may read stale files, observe a service that has not reloaded the right code, write artifacts, or return an observation whose equality is wrapper-specific. The systems question is therefore not whether a future call can be guessed, but when a result already produced by speculative execution can safely become task-visible.

The design draws on Tomasulo's algorithm~\cite{tomasulo1967efficient,lipasti1996exceeding,smith1988precise}, as shown in the CPU analogy of Figure~\ref{fig:intro-overview}(c): issue in order, execute as soon as operands are ready, and commit in order so that visible state is never speculative. \system applies the same separation to tool calls: a tool action plays a long-latency instruction, committed task state plays architectural state, and in-order commit keeps speculative state invisible. Tool actions, unlike instructions, are generated rather than fetched, their read sets are undeclared, their observations are wrapper-canonical rather than bit-exact, and their effects land in files and services rather than registers. This is why \system distinguishes \textit{ready operands}, such as workspace file versions copied into an overlay and revalidated at commit, from \textit{in-flight operands}, such as a loaded-version record that only a service-restart action can produce. A call whose dependencies or effects cannot be captured is a \textit{speculation barrier} and stays on the serial path.

Existing systems stop at narrower reuse boundaries. Multi-call parallel dispatch~\cite{kim2024llmcompiler} overlaps only actions already emitted in the same turn. PASTE, SPORK, and related tool-call speculation systems predict future calls but can safely keep only side-effect-free results or reuse at the action/output boundary~\cite{sui2026paste,bai2026spork,ji2026speculate}. \system instead validates at the dependency and effect boundary: a drafted tool action may run out of trajectory order and mutate its overlay, but it commits only as part of the \textit{validated action prefix}: the longest run of consecutive trajectory steps, starting at the next uncommitted one, whose Trace IR records all pass commit validation.

We evaluate \system on three benchmarks spanning sub-second tool calls to
hours-long coding sessions, using two H20 GPUs. \system achieves a
1.31$\times$ speedup on 100 sampled SWE-bench Verified tasks, 1.35$\times$ on
28 sampled Terminal-Bench~2.0 tasks, and 1.27$\times$ matched progress across
18 SWE-Marathon tasks. Across 4{,}010 audited commit-validation records, it
produces no false accepts.

This paper makes the following contributions:

\begin{itemize}
\item A technique for accelerating agent tool calls through out-of-order
speculative execution and selective re-execution.
\item The design and implementation of \system, which combines speculative
tool-call execution with dependency analysis to accelerate long-horizon agent
workloads without modifying applications.
\item An evaluation of \system demonstrating its feasibility and generality
across multiple real-world use cases.
\end{itemize}

\section{Agent Harness Background}
\label{sec:motivation}

\subsection{Execution Model}

An \textit{agent harness} is the runtime between an LLM and its environment: it builds prompts, invokes the model, executes the tool calls in each response, and appends their results to the conversation~\cite{yao2023react,schick2023toolformer}.
An \textit{action} is one tool call; its \textit{observation} is the result returned to the model; a \textit{turn} groups one LLM response with the actions and observations it triggers; and a \textit{trajectory} is the complete sequence for one task.

At turn $i$ the harness supplies the conversation and current workspace to the model, which emits a batch $B_i$ of actions.  The harness executes them, collects observations, and only then does the model produce turn $i{+}1$.  Because actions within a single batch may run concurrently, loop latency is approximately
\[
T_{\mathit{loop}}
=
\sum_i \left(T_{\mathit{llm}}(i)
+ \max_{a \in B_i} T_{\mathit{tool}}(a)\right).
\]
The summation captures the \emph{observation boundary} between turns: a later-turn action does not yet exist at that boundary, so ordinary multi-call dispatch cannot start it. Cross-turn speculation instead creates a \textit{candidate}, a predicted future action whose execution and effects remain invisible until the main agent emits the matching canonical action and the runtime validates its result.

\subsection{Cross-Turn Opportunity}

Section~\ref{sec:introduction} establishes the scale of the observation stall; the reusable dependence slack hidden inside those stalls is the \textit{cross-turn slack} that \system targets. The remaining question is whether that slack is large enough to exploit. The wait is concentrated rather than spread: an average SWE-Marathon trajectory contains 31 actions lasting at least 30\,seconds, clone-style tasks almost none, and refactor- or training-style tasks 22--75. Since Section~\ref{sec:eval-overheads} measures a per-candidate break-even of roughly one to twelve seconds of tool time, depending on the fork path, \system targets long trajectories where a few external observations repeatedly create critical-path stalls.

\begin{figure}[!t]
  \centering
  \includegraphics{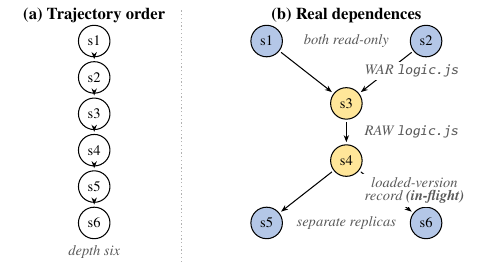}
  \caption{Trajectory order versus true dependences for six consecutive calls from a measured \texttt{slack-clone} run. The dependence DAG reduces the serial chain of depth six to a critical path of depth four.}
  \Description{Two panels. The left panel shows six circles labelled s1 through s6 joined top to bottom by arrows into a single chain, labelled depth six. The right panel shows the same six calls as a graph: s1 and s2 sit side by side with no edge between them and both point to s3; s3 points to s4; and s4 points to both s5 and s6, which sit side by side with no edge between them. Edges carry labels reading write-after-read (WAR) on logic.js, read-after-write (RAW) on logic.js, and loaded-version record.}
  \label{fig:motiv-dag}
\end{figure}

Figure~\ref{fig:motiv-dag} makes this concrete on six consecutive calls from a measured \texttt{slack-clone} run~\cite{swemarathon2026}. In trajectory order the calls form a chain of six, but their actual read and write sets reduce it to a critical path of four: trajectory order is not execution order. The stall hides work that a serial harness cannot issue because the later-turn actions have not yet been decoded, not because they are truly dependent.

Multi-call dispatch cannot exploit this slack: fewer than 10\% of turns emit two or more actions, and published measurements report similarly limited intra-turn parallelism in GAIA and SWE-bench~\cite{bai2026spork,mialon2023gaia,jimenez2024swebench}. Section~\ref{sec:design-overview} shows how \system turns this cross-turn opportunity into safe speculative execution.

\section{TomasuLLM Design}
\label{sec:design}
\label{sec:design-overview}%
\label{sec:isolated-commit}%
\label{sec:dep-tracking}%

\system drafts future actions, runs them in isolated copy-on-write sandboxes, traces their dependencies and effects, and commits results in trajectory order only after validation against committed state. The main agent remains the sole authority: speculative work can be discarded, but nothing it publishes can diverge from what serial execution would produce.

The central mechanism is an \textit{operand split}. We adopt the term \textit{operand} by analogy with out-of-order processors; the entities tracked are file versions and process-state records, not register values or memory addresses. Every operand \system can identify falls into exactly two classes: a \textit{ready operand} that already exists in the workspace and can be copied, or an \textit{in-flight operand} still inside an uncommitted producer. The first can be read early and validated at commit; the second forces a wait. This split---not the tool type, not the action---is what determines whether a call can run ahead, and everything else in the runtime exists to make that split safe to act on. A call whose read set or effects cannot be captured offers no operands to classify and stays on the serial path as a \textit{speculation barrier}.

\begin{figure}[!t]
    \centering
    \includegraphics[width=\columnwidth]{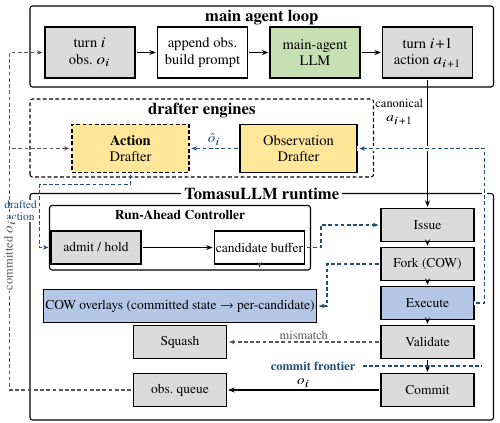}
    \caption{\system runtime with drafter engines.}
    \Description{A single-column block diagram with three stacked boundaries. The top row is the main-agent loop, with four equal boxes for the committed observation, prompt construction, main-agent LLM, and next action. The middle row is drafter engines, with equal boxes for the Action Drafter and Observation Drafter connected by a predicted-observation arrow. The bottom row is the TomasuLLM runtime. Its right side is an equal-size vertical pipeline from Issue, Fork, Execute, Trace IR record, Validate, to Commit, with a commit frontier before Commit. Its left side contains compact equal-size boxes for the Run-Ahead Controller and COW overlays, plus Squash and the committed-observation queue. Straight arrows show the canonical action entering Issue, drafted actions entering the controller, candidate execution feeding the Observation Drafter, and validated observations returning to the main-agent loop.}
    \label{fig:design-overview}
\end{figure}

The design takes its shape from Tomasulo's algorithm: issue in order, execute as soon as operands are ready, and commit in order so that visible state is never speculative. The two operand classes map to the two a core already distinguishes: a workspace file version plays a role analogous to a renamed register---bound at read and squashed if stale at commit, with the COW overlay serving as the private copy rather than a hardware register pool; a loaded-version record plays a role analogous to an operand not yet produced, so \texttt{restart}$\rightarrow$\texttt{test} is a true dependence analogous to store$\rightarrow$load. The analogy stops at content: an out-of-order core speculates on control flow over a fixed instruction stream, whereas the Action Drafter generates an action's content, so a misprediction can be an action that never occurs and commit validation must check the action itself.

Figure~\ref{fig:design-overview} shows the runtime. The main-agent loop (top) produces a canonical action $a_{i+1}$, which enters the runtime at Issue. In parallel, the Action Drafter proposes a candidate from committed history; the Observation Drafter predicts its result $\hat{o}_i$ so the next draft can start without waiting. Inside the runtime, the Run-Ahead Controller admits or holds each candidate, Fork creates a private copy-on-write overlay, and Execute runs the tool. Validate checks the Trace IR record: a mismatch triggers Squash; a pass advances the commit frontier and returns the validated observation to the main-agent loop through the observation queue.

A call \textit{issues} non-speculatively when it is the canonical action decoded by the main-agent LLM, or speculatively when it is a candidate drafted ahead of the frontier. The runtime sits below an unchanged harness, between the point where a call is parsed and the point where its observation is appended (Section~\ref{sec:impl-harness}); the policy, prompts, tool schemas, and decoding are unchanged. A \textit{tool wrapper}, the per-tool code that executes one call and formats the observation the model sees, exposes the dependency and effect evidence that validation needs.

\begin{figure*}[t]
    \centering
    \includegraphics[width=\textwidth]{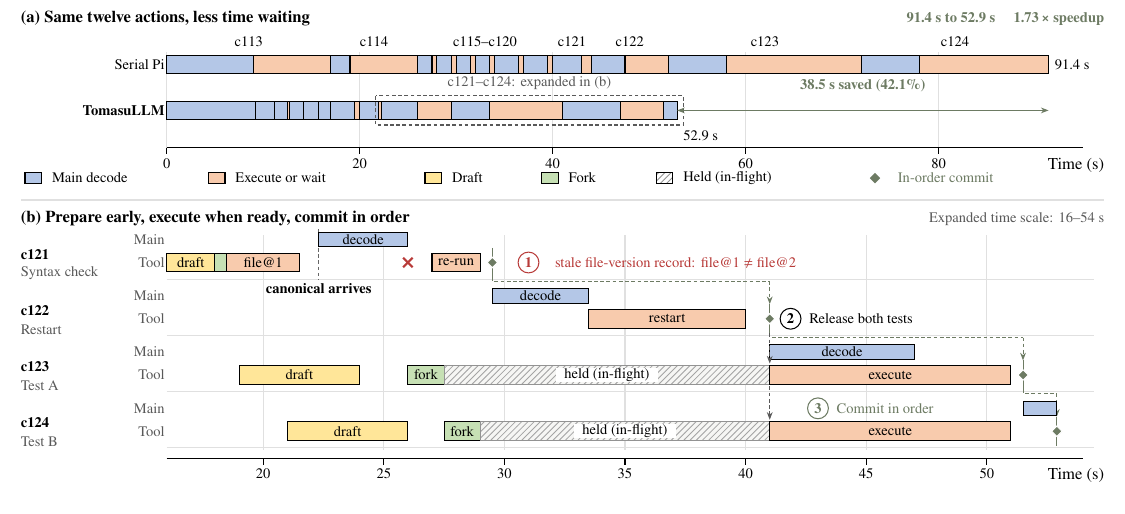}
    \caption{Walkthrough trace for twelve \texttt{slack-clone} actions. (a)~Serial vs.\ run-ahead timing; (b)~c121--c124 expanded, showing stale-file rejection, waiting on c122's loaded-version record, and in-order commit.}
    \Description{Two panels. Panel a shows two horizontal bars on one time axis from 0 to 92 seconds: the serial-baseline bar alternates blue decode segments with orange tool-wait segments for twelve actions and ends at 91.4 seconds; the TomasuLLM bar is mostly blue with three orange segments and ends at 52.9 seconds, with a dashed box around 21 to 54 seconds and a red cross at 26 seconds. Panel b enlarges the box as four rows, c121 to c124: c121 shows draft, fork, an early execution labelled file at version one, a red cross, a re-run segment and a decode strip above; c122 shows a decode strip and a restart bar with a commit mark at 41 seconds from which dashed arrows descend to c123 and c124; c123 and c124 each show draft, fork, a hatched held segment ending at 41 seconds, an execute bar to 51 seconds and a decode strip that starts after execution began; commit marks appear in order. Circled numbers one, two and three mark the three cases.}
    \label{fig:eval-trace}
\end{figure*}

\subsection{Drafting Pipeline and Speculation Barriers}

\label{sec:scheduling}

\system's pipeline has two prediction steps. The \textit{Action Drafter} proposes a concrete next tool action from the current or predicted history. As soon as that candidate is issued, and without waiting for it to return, the \textit{Observation Drafter} predicts the model-visible observation it will produce, and that predicted observation becomes the parent context for the next Action Drafter call.

Not waiting is the point: a chain can be drafted while its first link is still executing, which is what makes cross-turn slack reachable at all. The controller repeats the draft-and-predict step up to chain depth $K$ or until the candidate budget is full. Live candidates occupy the \textit{candidate buffer}: a slot is taken when a candidate issues and released when it commits or is discarded, so the candidate budget is the number of slots.

The prediction is never evidence for the commit decision: $\hat{o}_i$ lets a candidate keep reasoning ahead, but the candidate commits only after sandbox execution produces a real $\tilde{o}_i$ and the validator accepts its Trace IR record (Section~\ref{sec:commit}). A call whose wrapper cannot define a deterministic observation, whose effects are irreversible, or whose dependencies cannot be traced conservatively offers no operands to classify and is therefore a \textit{speculation barrier} that runs serially. Classes that act as barriers include drafted edits (whose content depends on what the agent has read), service-restart actions, checkpoints, and final submissions.

\subsection{The Operand Rule and Scheduling}
\label{sec:operand-rule}

Two one-version policies fail in opposite directions. Treating every operand as ready runs a test before the edit and restart it was meant to check, so the validator rejects it and its execution time is paid again serially. Invalidating every pending test on any file write collapses the speculation window in exactly the edit--test loops that hold the most latency. The \textit{operand rule} sits between them: a tool call may run ahead exactly when every operand it reads already exists, and must wait exactly on the youngest producer of one that does not.

A candidate executes immediately when all operands are ready. If it names an in-flight operand, the controller still forks and prepares it, because building an overlay depends on the drafted tool action and its parent lineage rather than on the observation the tool will return. Only execution waits. A chained candidate follows the same rule applied to its speculative parent: it forks from the parent and runs at once when the parent writes nothing it reads, and otherwise forks only after the parent's execution finishes. When ready candidates outnumber execution slots, dispatch-order weights rank long tests above short tests and above reads; the weights change throughput, never what commits (Appendix~\ref{app:sched-settings}). The out-of-order execution window is therefore the set of candidates whose operands are all ready: within this window, execution order is determined by dispatch weights rather than issue order. A candidate held on an in-flight operand re-enters the window only when its producer commits, so the operand rule simultaneously enables out-of-order execution for ready candidates and enforces ordering for dependent ones.

The \textit{Run-Ahead Controller} uses \textit{source-limited scheduling}: every input about a future action must name an online source---the current execution history, tool schemas, and drafter outputs---rather than a static model of the task's future trajectory. Because commit validation alone decides publication, a bad scheduling choice can waste an overlay or miss an opportunity but never publish a wrong result. Section~\ref{sec:impl-sched} lists the online sources and Table~\ref{tab:sched-inputs} states what each one costs when it is wrong.

\begin{figure}[t]
\centering
\includegraphics{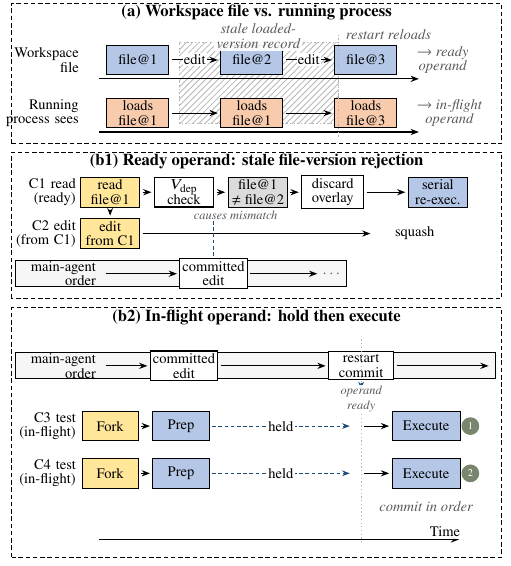}
\caption{Ready and in-flight operands. (a)~Workspace files track edits; the running process follows its loaded-version record. (b1)~A ready read runs early; a committed edit triggers $V_{\mathrm{dep}}$ mismatch and discards the overlay. (b2)~An in-flight dependent forks early but is held until the restart commits.}
\Description{Three stacked schematic panels. Panel a shows an upper workspace track stepping from file at version 1 to version 2 to version 3 at two edits, and a lower process track showing that the running process still loads file version 1 until a restart reloads file version 3. A shaded band marks the stale interval. Panel b1 shows Candidate 1 reading file at version 1, a V-dep check detecting version mismatch caused by the committed edit, discarding the overlay, and serial re-execution; Candidate 2 is chained on C1 and squashed. A causal arrow connects the committed edit to the mismatch. Panel b2 shows Candidates 3 and 4 forking and preparing early, held until the restart commit makes the loaded-version record ready, then executing; in-order commit marks appear.}
\label{fig:execution-snapshots}
\end{figure}

Figure~\ref{fig:eval-trace} traces twelve consecutive actions from the SWE-Marathon \texttt{slack-clone} task and introduces the three cases used throughout this section: a stale rejection, a held-then-executed candidate, and in-order commit. The stale candidate runs early and loses because a file-version record changes before the frontier reaches it. The held tests fork and prepare early, then execute only after the service-restart action publishes the loaded-version record they observe. The commit marks remain in main-agent order even when execution finishes out of order.

Figure~\ref{fig:execution-snapshots} uses the running trace to show the dependence problem. In the figure, file@$k$ names the $k$th committed version of one workspace file, the lower track shows the running process's loaded-version record, and C1--C4 denote speculative candidates. Panel~(b1) shows the ready-operand case: C1 reads an existing file version and C2 chains from it, but the committed edit changes the version, so $V_{\mathrm{dep}}$ rejects C1 and C2 is squashed. Panel~(b2) shows the in-flight case: C3 and C4 are tests that fork early but are held until the restart commits.

\subsection{Ready Operands: Workspace Files}
\label{sec:versioning}

A workspace file exists when a candidate is drafted, so \system copies its content into the candidate's overlay and lets the call execute early. The \texttt{read} in our trace is drafted while the preceding action is still running, and it runs at once. A copied file can still go stale if the main agent's edit overwrites it before the candidate reaches the frontier. \system therefore records the version of every file an execution read (Section~\ref{sec:impl-tracing}) and rechecks those versions against committed state at commit. A mismatch discards the overlay and sends the call to serial re-execution, as for candidate C1 in Figure~\ref{fig:execution-snapshots}(b1) and, in Figure~\ref{fig:eval-trace}, for the syntax check c121, whose copy of \texttt{src/logic.js} the committed edit replaces before the frontier reaches it. A ready operand never forces a wait; the cost of being wrong is one discarded overlay and one serial re-execution. Reads and syntax checks are handled this way, as is a \textit{canonical} \texttt{edit} the main agent has actually issued, which executes in an overlay like any other call. (A \textit{drafted} edit, by contrast, is a speculation barrier because its content is invented; see Section~\ref{sec:scheduling}.)

\subsection{In-Flight Operands: Producer Outputs}

An operand is \textit{in flight} when an older action has yet to produce it and that action cannot run speculatively. A coding agent has two producer classes, producing at different levels, and only one of them leaves an operand in flight. An \texttt{edit} writes content that depends on what the agent has read, so a drafted edit is invented text and is never dispatched; what it produces is a new file version, which every younger call still consumes as a ready operand, present at some version, copied, and rechecked at commit. A service-restart action produces the other level: a COW overlay can hold file changes, but a later test observes the shared process only after it reloads committed code, and until that reload the loaded-version record the test needs exists nowhere to be copied. Only this second level forces a wait. The second \texttt{test}, drafted after the edit and before the restart, is therefore held until the service-restart action commits.

Nothing about a held test's execution moves early; what moves is the decision and the overlay fork, which a serial runtime performs only after the producer commits. For each contacted process, \system adds its loaded-version record to the test read set (Section~\ref{sec:impl-sandbox}); a test binds to it only when its own arguments declare it runs against the shared process, with the most recent preceding restart as producer.

\subsection{Isolated Execution}
\label{sec:spec-pipeline}

Each candidate executes inside a private copy-on-write overlay forked from the committed state or, for a chained candidate, from its speculative parent. Writes stay inside the overlay until commit; effects that cannot be deferred, such as network sends, are blocked before execution (Section~\ref{sec:impl-sandbox}). The overlay is the unit of both isolation and recovery: a failed validation discards the overlay and its contents, so no speculative state leaks. Sibling candidates cannot conflict: each writes only inside its own overlay, and a drafted edit is never dispatched, so the only writes that cross candidates come from an edit the main agent has already issued, which commit in order.

\system applies precise-state discipline~\cite{smith1988precise} at observation boundaries, but the goal is correctness of the task state rather than replaying a baseline trajectory. \textit{Response reuse} is reuse, not prediction: the Action Drafter never substitutes for main-agent decoding, and response text is kept only when the same main-agent checkpoint has already produced it under a replayable decoding state and the corresponding action and tool-result records validate at the frontier.

\subsection{Commit Validation}
\label{sec:trace-ir}%
\label{sec:commit}%
\label{sec:commit-correctness}%

When the \texttt{read} of our trace finishes inside its overlay, the main agent has not yet asked for it. The validator will later have to answer four questions without executing the tool again: did the candidate run the action the main agent now wants, is what it read still current, does its observation canonicalize under the wrapper's rule, and do the effects it would publish still apply. These are the four links of a tool call's causal chain---action, dependencies, observation, effects---and each can diverge from what serial execution would produce. The Trace IR record a sandboxed execution leaves behind answers all four. Appendix~\ref{app:trace-ir-schema} maps each field to its check and shows the record one measured call left behind in Figure~\ref{fig:trace-ir-example}. A predicted observation appears nowhere in the record.

A read set cannot be declared, because a \texttt{bash} command hides what it reads inside an opaque process, so it must be traced conservatively enough that no dependency is missed. \system embeds Riker~\cite{curtsinger2022riker} inside each E2B sandbox overlay, inheriting its conservative syscall-tracing guarantee: every file read, write, and absence check the sandboxed process tree performs is recorded, and any access that falls outside the traced tree makes the call a speculation barrier (Appendix~\ref{app:tracing}). Section~\ref{sec:impl-tracing} describes how each Trace IR field is captured. Deferrable effects are the one addition, following hS, a speculative executor for shell scripts~\cite{hs2026}.

Commit is the only point where speculative work becomes visible. \system's validator advances the \textit{commit frontier}---the validated action prefix introduced in Section~\ref{sec:introduction}---through the longest run of consecutive actions whose records validate in trajectory order. For candidate $i$, reusing a sandboxed tool result requires
\[
V_{\mathrm{reuse}}(i)
=
V_{\mathrm{act}}
\land
V_{\mathrm{dep}}
\land
V_{\mathrm{record}}
\land
V_{\mathrm{effect}} ,
\]
one predicate per link in the causal chain. The four checks rest on the same determinism assumption as Riker: a tool's observation and effects are a function of what it read, provided the read set is traced conservatively, so a record whose action, reads, observation, and effects all validate is one the tool would reproduce if run from committed state now.

\paragraph{$V_{\mathrm{act}}$: Action identity.}
Checks that the canonical action equals the main agent's next action at the frontier.

\paragraph{$V_{\mathrm{dep}}$: Dependency freshness.}
Rechecks recorded reads and absences against committed file versions and each loaded-version record. A recorded read is meaningful only relative to the parent state the candidate forked from, so $V_{\mathrm{dep}}$ first checks that recorded lineage against the frontier and rejects a candidate forked from a superseded or squashed parent before comparing any digest. In the walkthrough, $V_{\mathrm{dep}}$ catches c121's stale read of \texttt{src/logic.js} and the call re-executes serially.

\paragraph{$V_{\mathrm{record}}$: Observation integrity.}
Checks that the sandbox observation carries the content and digest required by the wrapper's canonicalization rule. The wrapper strips non-deterministic fields so the observation is a deterministic function of what the tool read.

\paragraph{$V_{\mathrm{effect}}$: Effect safety.}
Checks that the recorded effect deltas can still be promoted or replayed (Appendix~\ref{app:promotion}) and that their post-state digests still hold.

\paragraph{Completeness.}
The four checks cover the complete causal chain: $V_{\mathrm{act}}$ ensures the same input, $V_{\mathrm{dep}}$ the same preconditions, $V_{\mathrm{record}}$ the same wrapper-canonical output, $V_{\mathrm{effect}}$ the same captured side effects. Together, the speculative execution is indistinguishable from serial execution within the model-visible wrapper scope. Appendix~\ref{app:trace-ir-schema} gives the inductive argument: the base case $n{=}0$ is trivial, and each candidate passing $V_{\mathrm{reuse}}$ extends the validated action prefix by one step.

Commit then takes one of two paths. When the candidate's overlay descends directly from the frontier, the overlay itself becomes committed state; otherwise the recorded effect deltas are replayed onto committed state (Appendix~\ref{app:promotion}).

\subsection{Prediction Consistency and Squash}
\label{sec:squash}

A separate predicate, $V_{\mathrm{pred}}(i) = [\,\hat{o}_i = \tilde{o}_i\,]$, decides only whether speculative work that consumed the predicted observation $\hat{o}_i$ stays eligible. It is not part of $V_{\mathrm{reuse}}(i)$: after $V_{\mathrm{reuse}}(i)$ succeeds, candidate $i$ can commit on its own real execution even when $\hat{o}_i \ne \tilde{o}_i$; the mismatch squashes only the descendants drafted from $\hat{o}_i$. It never exempts a descendant from that descendant's own checks.

A candidate chain carries two dependences: the younger runs in the older's overlay, and it was drafted on the older's predicted observation. At the frontier, a passing candidate publishes effect deltas and advances; a failing candidate is discarded and re-executed serially; and a $V_{\mathrm{pred}}$ mismatch after $V_{\mathrm{reuse}}$ succeeds commits the current candidate but squashes descendants drafted from the wrong prediction.

\subsection{Run-Ahead Bounds}
\label{sec:run-ahead-bounds}

The admission rule follows the pipeline of Figure~\ref{fig:design-overview}. The controller chooses only among drafted tool actions emitted online, admits a candidate only when its tool class and resources permit speculation, forks it at once and holds it only on an in-flight operand, and draws the next draft from a predicted observation produced concurrently with execution. Nothing on that path settles correctness: $V_{\mathrm{reuse}}$ and $V_{\mathrm{pred}}$ are both evaluated at the frontier, so the main agent may later emit a different canonical action, in which case the candidate fails $V_{\mathrm{act}}$ and is discarded. This is the point at which the runtime enforces that scheduling uses no oracle trajectory.

Run-ahead is bounded on three coupled axes. \emph{Width}---the candidate budget, forks in progress, and execution slots---is an allocation; \emph{depth} is bounded by $K$ at draft time and by the first failed predicted observation at commit time; and the \emph{structural} bound, the operand rule and the speculation barriers, acts on the other two: the distance from the frontier to the next producer, not $K$, sets how much of a chain can run. Section~\ref{sec:eval-roofline} reports the resulting dependence roofline.

\section{Implementation}
\label{sec:implementation}

The \system prototype comprises approximately 11,200 lines of TypeScript across the Run-Ahead Controller, Pi integration, E2B overlay management, traced tools, canonicalization, and Trace IR validation. It is implemented on Pi v0.84.2~\cite{earendilpi2025}, an open-source coding-agent harness. Setting chain depth to zero yields \textit{Serial Pi}, the base agent of Figure~\ref{fig:intro-overview}(a), with the same main-agent configuration and tools but no speculative run-ahead; Serial Pi is the serial baseline throughout Section~\ref{sec:evaluation}.

\paragraph{Harness integration.}
\label{sec:impl-harness}
The two interposition points of Section~\ref{sec:design-overview} sit at Pi's tool-execution boundary: the dispatch side sees a parsed action and matches it against the candidate buffer, launches a candidate, or takes the serial path; the observation side sees the tool result before it is appended to the model context and runs commit validation there, returning either a validated observation or a serial re-execution result. Both sides speak a small action/observation record (Appendix~\ref{app:atif}), so porting \system to another harness needs a new adapter and compatible tool wrappers, not a new protocol.

\paragraph{Overlays and process loads.}
\label{sec:impl-sandbox}
E2B~2.39.1 implements sandbox cloning, promotion, release, and retry handling~\cite{e2b2026snapshots}. Each task runs in one committed root sandbox instrumented with Riker. E2B snapshots implement COW overlays, taken from the committed parent or, for a chained candidate, from the speculative parent, together with a private copy of any local state the call would mutate. A read-only call forks in about 0.08\,s and a test that needs a private data copy in 3--5.5\,s; Table~\ref{tab:primitive-costs} reports the primitive measurements. The tool runs inside the overlay and leaves its Trace IR there.

The \texttt{restart} wrapper implements service-restart actions. The server process under test is a shared process in the root sandbox; the wrapper reloads it from the committed workspace, waits for its readiness probe, and records the loaded-version record. Riker's tracing stops at the wrapper's process tree, so the shared process is covered by this record rather than traced into. A candidate may depend on the shared process only through that record; a call that cannot prove it leaves the process's in-memory state unchanged is a speculation barrier. Appendix~\ref{app:process-loads} gives the journal evidence for hold and release events.

\paragraph{Capturing Trace IR.}
\label{sec:impl-tracing}
\system runs Riker~\cite{curtsinger2022riker} inside each overlay. Transparent wrappers derive dependencies from call arguments and state digests; an opaque command runs under \texttt{rkr}, and \system lowers the file events Riker records into the read, absence, write, and effect sets of that call's Trace IR record (Appendix~\ref{app:tracing}). Coverage is the local process tree Riker traces, ordinary \texttt{fork} and \texttt{exec} children included. A command whose dependencies or effects the record cannot prove complete never reuses a result (Appendix~\ref{app:wrapper-schema}). Durable journals record each validation predicate and whether a result was promoted, replayed, rejected, or serially re-executed; these records drive the validation audit in Section~\ref{sec:evaluation}.

\paragraph{Commit, promotion, and replay.}
\label{sec:impl-commit}
Appendix~\ref{app:promotion} gives the promotion, replay, and squash paths of Section~\ref{sec:commit} at implementation level.

\paragraph{Drafters and runtime settings.}
\label{sec:impl-sched}
Serial Pi and the \system main agent use the same Qwen3.8-27B checkpoint, prompt, tool schema, tool wrappers, decoding configuration, and completion rule. The Action Drafter is Qwen3-8B with a LoRA adapter fine-tuned on SWE-Marathon tool-call trajectories; the Observation Drafter uses the off-the-shelf 35B Qwen-AgentWorld checkpoint without adaptation~\cite{zuo2026qwenagentworld}. The runtime serves both through independently configured OpenAI-compatible endpoints. Appendix~\ref{app:drafter-params} gives the corpus, split, masking procedure, adapter checkpoint, hyperparameters, and drafting latencies.

The prototype uses only the online inputs of Section~\ref{sec:scheduling}. The Run-Ahead Controller admits only candidates whose drafted tool action passes the tool registry's deferrable-effect and traceability checks; a bad draft fails $V_{\mathrm{act}}$ at the frontier and costs one discarded overlay. Resource limits appear in Appendix~\ref{app:sched-settings}.

\section{Evaluation}
\label{sec:evaluation}
\label{s:eval}

\newcommand{\evalplot}[1]{%
  \includegraphics[width=\linewidth]{figures/#1.pdf}%
  \vspace{-0.45em}%
}
\newlength{\evalFigureSevenHeight}
\setlength{\evalFigureSevenHeight}{0.422\linewidth}
\newcommand{\evalplotfiguresevenheight}[1]{%
  \includegraphics[height=\evalFigureSevenHeight]{figures/#1.pdf}%
  \vspace{-0.45em}%
}

\begin{figure*}[t]
  \centering
  \includegraphics[width=\linewidth]{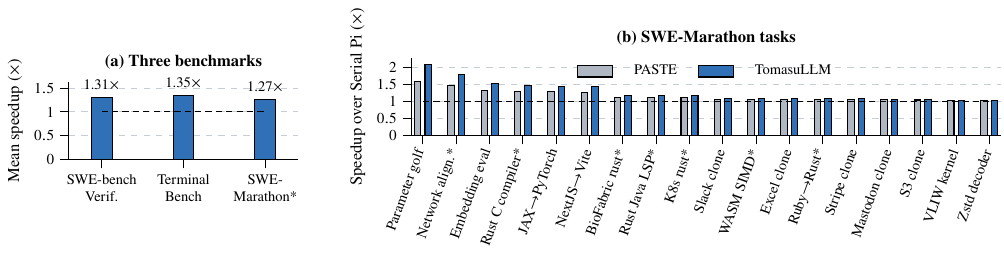}%
  \vspace{-0.45em}%
  \caption{End-to-end speedup over Serial Pi. Panel (a) reports the three benchmark
  means; panel (b) reports per-task SWE-Marathon speedup for Prefetch (PASTE)
  and \system. Starred entries report matched-progress time at the 6.0\,h E2B
  cap rather than task completion.}
  \Description{Two-panel speedup figure. The left panel shows mean speedup
  for SWE-bench Verified, Terminal-Bench 2.0, and SWE-Marathon. The right panel
  shows grouped bars of per-task speedup over Serial Pi for the PASTE-style
  prefetch baseline and \system across 18 SWE-Marathon tasks, sorted by \system
  speedup. Starred labels denote matched-progress rows at the 6.0 hour cap;
  dashed lines mark the Serial Pi baseline.}
  \label{fig:eval-speedup}
  \end{figure*}
Our evaluation answers the following research questions:

\begin{enumerate}[label=\underline{\emph{Q\arabic*.}},leftmargin=*]
\item Can \system reduce end-to-end latency over strong baselines, and which
workload regimes benefit most?
(\S\ref{sec:exp_q1})
\item Does commit validation preserve task-execution correctness and task
outcomes under speculative execution?
(\S\ref{sec:eval-correctness})
\item Does the speedup come from \system's design rather than ordinary
concurrent execution or one-step prefetching?
(\S\ref{sec:exp_q2})
\item What runtime costs and structural limits bound the speedup, and when does
speculation break even?
(\S\ref{sec:exp_q3})
\end{enumerate}

\inlinesection{Experimental environment.}
Serial Pi and \system share the configuration of
Section~\ref{sec:impl-sched}. The three model engines run with vLLM in FP8 on two H20 GPUs: the main agent and
Action Drafter share one card, and the Observation Drafter uses the second.
Tool calls run in CPU-only E2B sandboxes on a 2$\times$ Intel Xeon Gold 6548Y+
server with 1.34\,TiB RAM; all latencies include sandbox setup, teardown, and
main-agent/drafter contention. By default, experiments use chain depth six,
three live candidates, at most two concurrent overlay forks, and eight execution
slots. Section~\ref{sec:implementation} gives the implementation details and
Appendix~\ref{app:drafter-params} gives full drafter settings.

\subsection{End-to-End Latency}
\label{sec:exp_q1}
\label{sec:eval-setup}
\label{sec:eval-e2e}

We evaluate \system on 18 SWE-Marathon tasks that fit the CPU-only sandbox,
covering 301--7{,}677 recorded tool actions and excluding the benchmark's two
GPU-dependent tasks. Eleven complete under both \system and Serial Pi; on the
seven timeout-capped tasks, the 6.0\,h entry is the E2B cap rather than a
completion time. Rather than requiring task
completion, we anchor on the number of committed tool results and report a
\textit{matched-progress ratio}: 6.0\,h divided by
the time a configuration takes to commit as many tool results as Serial~Pi
has committed at the cap. To cover shorter tool calls, we also
evaluate 100 SWE-bench Verified
tasks~\cite{jimenez2024swebench,openai2024swebenchverified} and 28
Terminal-Bench~2.0 tasks~\cite{tb2}, sampled at random from its 89 tasks with
a fixed seed. Each configuration runs three fresh online executions per
task, and we aggregate per-task medians with the geometric mean to reduce
run-to-run noise.

For the real SWE-Marathon wall-clock runs, we compare three configurations
while holding the harness, main model, sandbox backend, Action Drafter, and
workload fixed. \emph{Serial Pi} disables \system. \emph{Prefetch (PASTE)}
follows the one-step speculation boundary of PASTE~\cite{sui2026paste}
(reimplemented; no open-source artifact is available): it predicts and
pre-executes only the next side-effect-free call, using the same Action
Drafter and sandbox as \system but disabling observation prediction, Trace IR
evidence, operand-aware run-ahead, and selective replay. \emph{\system} adds
those mechanisms with in-order commit.
Figure~\ref{fig:eval-speedup} panel~(b) sorts tasks by \system speedup.

Table~\ref{tab:eval-wallclock} gives the corresponding SWE-Marathon wall-clock
times. Figure~\ref{fig:eval-speedup} reports the SWE-Marathon aggregate as a
1.27$\times$ mean matched-progress speedup (arithmetic mean over the 18
tasks); the star marks rows where Serial Pi
hits the E2B cap and \system is timed at the same progress point. Across the
eleven non-capped entries, \system reaches a 1.23$\times$ geometric-mean
speedup over Serial Pi. On the seven capped rows, the matched-progress
geometric mean is 1.26$\times$. The per-task speedup ranges
from 1.03$\times$ to 2.07$\times$. The strongest rows are the tool-dominated cases:
Parameter golf reaches 2.07$\times$, Network alignment reaches 1.78$\times$,
Embedding eval reaches 1.51$\times$, and Rust C compiler reaches 1.47$\times$.
Prefetch (PASTE) reaches 1.16$\times$ over all 18 rows; \system is
faster on every row because it can validate and reuse cross-turn tool
executions when the Trace IR record still matches committed state.

\begin{figure}[t]
\centering
\includegraphics[width=\columnwidth]{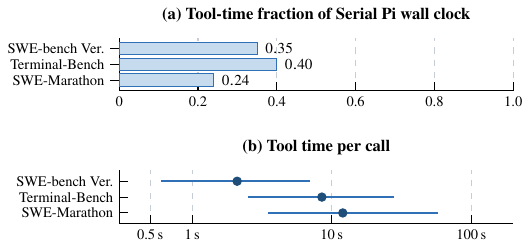}
\caption{Tool time on the Serial Pi baseline. (a) Share of end-to-end wall
clock spent inside tool executions; the SWE-Marathon bar averages the per-task
shares of Table~\ref{tab:eval-wallclock}. (b) Median and interquartile range of tool
time per call.}
\Description{Two stacked panels. Panel a is a horizontal bar chart of tool-time fraction with SWE-bench Verified at 0.35, Terminal-Bench 2.0 at 0.40 and SWE-Marathon at 0.24. Panel b plots tool time per call on a logarithmic axis, with median markers at 2.1, 8.5 and 12 seconds and interquartile ranges spanning 0.6 to 7, 2.5 to 28 and 3.5 to 58 seconds.}
\label{fig:eval-toolfrac}
\end{figure}

Figure~\ref{fig:eval-toolfrac} characterizes the three benchmarks on the Serial
Pi baseline. Tool executions occupy 0.35 of end-to-end wall clock on SWE-bench
Verified and 0.40 on Terminal-Bench~2.0. On SWE-Marathon the share varies by
more than an order of magnitude across tasks (Table~\ref{tab:eval-wallclock}):
from 3.6\% on Zstd decoder and 4.4\% on VLIW kernel to 62.7\% on Network
alignment and 74.0\% on Parameter golf, with a median of 12.0\% and a mean of
24.3\%. Because the main agent performs the same decoding work under both
configurations (Section~\ref{sec:eval-overheads}), a task whose Serial Pi run
spends a fraction $f$ of its wall clock in tools can gain at most $1/(1-f)$
from hiding tool time: 1.04$\times$ for Zstd decoder, 1.14$\times$ for Slack
clone, and 3.84$\times$ for Parameter golf. Per-call tool time decides whether
an individual candidate outlasts its own preparation: the median call takes
2.1\,s on SWE-bench Verified, 8.5\,s on Terminal-Bench~2.0, and 12\,s on
SWE-Marathon, where the upper quartile reaches 58\,s.
Accordingly, tasks whose Serial Pi run spends less than 15\% of wall clock in
tools average 1.07$\times$ speedup, while tasks above 40\% average 1.60$\times$.
Across the three benchmarks, SWE-bench Verified reaches 1.31$\times$ with
35\% tool time, Terminal-Bench~2.0 reaches 1.35$\times$ with 40\%, and
SWE-Marathon reaches 1.27$\times$ matched progress with 24\% mean tool time.
Median latency is no worse than Serial Pi; Appendix~\ref{app:eval-extra}
gives the per-task distribution.

\inlinesection{Generality across target models.}
The results above use Qwen3.8-27B as the target model. To test whether the
speedup depends on that choice, we repeat the 18 SWE-Marathon runs with
DeepSeek~V4~Pro as the main agent, holding the harness, tool wrappers,
sandbox backend, Observation Drafter, and scheduler settings fixed; only the
Action Drafter changes, re-fine-tuned on DeepSeek~V4~Pro trajectories with
the recipe of Appendix~\ref{app:drafter-params}, reaching a comparable
end-to-end acceptance rate (0.48 versus 0.42). Per-task speedups
(Table~\ref{tab:eval-target-model}) agree with the Qwen configuration within
0.03$\times$ on seventeen of eighteen tasks, both configurations reach the
same 1.24$\times$ geometric mean over the 18 rows, and the task ranking is
unchanged. The one exception is the most tool-dominated task (Parameter golf,
2.23$\times$ versus 2.07$\times$ at 74.0\% tool time), where the attainable
gain $1/(1-f)$ is most sensitive to run-to-run tool-time variance. The speedup therefore tracks each
task's tool-time structure rather than the identity of the target model, and
porting \system to a new target requires only re-fine-tuning the Action
Drafter on that model's trajectories.

\subsection{Correctness and Outcome Preservation}
\label{sec:eval-correctness}

Speculation must preserve two properties: \emph{commit-order integrity}
(no observation is published ahead of an older one) and
\emph{observation equivalence} (every committed result matches what
Serial~Pi would have produced at the same trajectory position).
We audit commit validation on SWE-Marathon, where tasks exercise
301--7{,}677 tool actions and provide the widest surface for false accepts.
We measure outcome preservation on SWE-bench Verified, where tasks complete
under both configurations and provide unambiguous pass/fail verdicts.

\begin{table}[t]
\centering
\caption{Fault injection. Hidden dependencies deliberately introduced into tool
wrappers, and the check that rejected each candidate.}
\label{tab:eval-fault}
\footnotesize
\begin{tabularx}{\columnwidth}{@{}Xccc@{}}
\toprule
Injected hidden dependency & Cases & Caught & First failing check \\
\midrule
Stale file version after committed edit & 4 & 4 & $V_{\mathrm{dep}}$ \\
Absence set: failed lookup, later create & 3 & 3 & $V_{\mathrm{dep}}$ \\
Directory rename changes path resolution & 3 & 3 & $V_{\mathrm{dep}}$ \\
Shared process state before restart & 3 & 3 & held on producer \\
Wall-clock or randomness in observation & 3 & 3 & $V_{\mathrm{record}}$ \\
Untraced read outside the process tree & 2 & 2 & speculation barrier \\
Non-idempotent network send & 2 & 2 & speculation barrier \\
\midrule
Total & 20 & 20 & 0 false accepts \\
\bottomrule
\end{tabularx}

\end{table}

\inlinesection{Commit order.}
Across the paired SWE-Marathon executions every committed call commits in
trajectory order, and no run publishes an observation ahead of an older one.

\inlinesection{Validation audit.}
We sample one call in ten across 18 paired SWE-Marathon executions,
yielding 4{,}010 commit-validation records: reads and writes pass
commit validation at 100\%, read-only \texttt{bash} calls at 99\%, and tests
at 98\%; the remainder are correctly rejected and re-executed serially.
An additional 390 barrier records run serially without reuse.
No false accepts occur across the full sample.

\inlinesection{Fault injection.}
The audit shows no false accept occurred; it does not show that one would have
been caught. Table~\ref{tab:eval-fault} closes the gap by injecting hidden
dependencies into tool wrappers and re-running the affected calls
speculatively. Twenty cases across seven classes all fail before commit---ten
at $V_{\mathrm{dep}}$, three at $V_{\mathrm{record}}$, three held on a
producer, and four blocked as speculation barriers. Read with
the audit, the 4{,}010 records place a 95\% upper bound of 0.075\% on the
false-accept rate.

\inlinesection{Outcome preservation.}
On the 100 SWE-bench Verified tasks (three runs each), both Serial Pi and
\system resolve a mean of 41.7\% (difference 0.0 points, 95\% CI
$[-4.3, +4.3]$); Serial Pi alone varies by four tasks across its runs, so
speculation stays within run-to-run noise. On SWE-Marathon, paired
hidden-verifier outcomes also match Serial Pi.

\subsection{Design Attribution}
\label{sec:exp_q2}
\label{sec:eval-mechanisms}

A natural objection is that the speedup comes from ordinary same-turn
concurrency or one-step prefetching rather than from \system's specific design.
We rule this out with single-mechanism ablations, a drafter-quality sweep, and
supporting diagnostics, all evaluated on 10 trajectories
(4{,}311 depth-1 canonical-action labels) reserved from the drafter
training corpus and never seen during training
(Appendix~\ref{app:drafter-params}); the end-to-end results in
\S\ref{sec:eval-e2e} use the full benchmark independently.

\begin{figure}[t]
\centering
\includegraphics[width=\columnwidth]{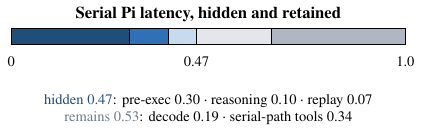}
\caption{Serial Pi latency normalized to one, split into the parts \system
hides and the parts that stay on the critical path.}
\Description{A stacked horizontal bar of Serial Pi latency normalized to one,
with tool pre-execution 0.30, retained reasoning 0.10 and replay 0.07 hidden,
and main-agent decode 0.19 and serial-path tools 0.34 remaining.}
\label{fig:eval-latency-breakdown}
\end{figure}

\paragraph{Each mechanism is individually necessary.}
We disable each of the four mechanisms in isolation and measure the increase in
Serial Pi's normalized latency
(Figure~\ref{fig:eval-latency-breakdown}). The full system hides 0.47 of the
baseline: 0.30 from early tool execution, 0.10 from retained downstream
reasoning, and 0.07 from Trace IR replay, with 0.05 added by discarded
suffixes, yielding a net normalized latency of 0.58.
Removing early tool execution alone adds 0.32; removing Trace IR evidence adds
0.18; removing operand-aware run-ahead adds 0.13; removing value speculation
adds 0.09. Even value speculation, the smallest contributor, still recovers
nearly twice the overhead it introduces.
The four deltas sum to 0.72, more than the 0.47 of hidden time, because the
mechanisms interact: early tool execution creates idle slots that Trace IR
replay and run-ahead then fill. Removing any one mechanism therefore also
removes its interactions with the others, so each delta over-credits the
removed mechanism. The key implication is the converse: \emph{no single
mechanism is redundant}---each one opens room that the others exploit.

\paragraph{Speedup is a predictable function of drafter quality.}
If speculative execution were an artifact of workload coincidence, replacing
the drafter with a weaker model should degrade speedup unpredictably. It does
not. A sweep over four LoRA checkpoints spanning a 20-percentage-point
acceptance range (Appendix~\ref{app:drafter-variant-sweep}) shows that
acceptance tracks action accuracy across all variants, and speedup is
linear in acceptance: each
10-percentage-point increase in acceptance rate lowers normalized latency by an
additional 0.047, with no sign of saturation. Extrapolating the fit places
break-even near an acceptance rate of 0.20. This linearity matters for two
reasons: it confirms that the verification pipeline adds negligible filtering
beyond the drafter's own prediction error, and it means that a better drafter
translates directly into faster end-to-end performance with no architectural
change.
$V_{\mathrm{act}}$ is empirically the binding constraint: the Action Drafter's
top-1 match is 0.46 and end-to-end acceptance is 0.42, a gap of only 0.04
from the other three checks combined. The Observation Drafter matches 65\% of
committed observations with an off-the-shelf model, leaving headroom on both
sides.

\begin{figure}[t]
\centering
\evalplot{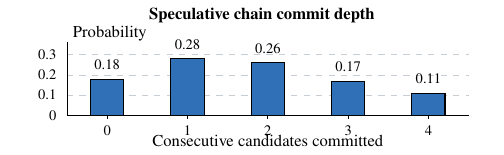}
\caption{Speculative chain diagnostics. (a)~Consecutive candidates committed before first chain termination. (b)~First mismatch cause share. (c)~Drafter match rates.}
\Description{Three panels. Panel~a is a bar chart giving the probability of each valid action prefix length from zero to four, with values 0.18, 0.28, 0.26, 0.17, and 0.11. Panel~b is a horizontal bar chart of first-mismatch shares: Action 0.36, Observation 0.26, Dependency 0.17, Speculation barrier 0.14, and Timeout 0.07. Panel~c is a horizontal bar chart of match rates: Action top-1 0.46, Action top-2 0.61, and Observation 0.65.}
\label{fig:eval-mechanisms}
\end{figure}

\paragraph{Supporting diagnostics.}
Two additional diagnostics corroborate the ablation.
First, speculative chains are shallow but effective
(Figure~\ref{fig:eval-mechanisms}): 82\% of chains commit at least one
candidate, yet no chain commits more than four. Depth is not the driver---a
single long test whose execution overlaps the main agent hides more wall time
than several correct sub-second reads, which is why early tool execution
dominates the ablation ranking.
Second, chain terminations map onto distinct design components: action mismatch
(36\%) and observation mismatch (26\%) are control and value mispredictions of
the two drafters; dependency mismatch (17\%) is caught by the dependency
checker against undeclared environment couplings; speculation barriers (14\%)
are static tool-registry annotations; timeout accounts for 7\%.
Notably, an observation mismatch does not reject the candidate---it commits the
candidate's own execution and squashes only descendants drafted from the wrong
predicted observation (Section~\ref{sec:squash}), preserving partial work.
Same-turn concurrency, which treats each invocation independently, has no
mechanism for such partial-chain commit.

\subsection{Costs and Limits}
\label{sec:exp_q3}
\label{sec:eval-overheads}
\label{sec:eval-roofline}
\label{sec:exp_q4}

Table~\ref{tab:primitive-costs} microbenchmarks the runtime primitives.
A candidate must recoup its preparation cost---overlay fork, trace recording,
dependency check, and commit (Table~\ref{tab:primitive-costs})---before it
saves wall time. Most candidates take the read-only fork path; under that
path the preparation overhead typically sums to 1--3\,s, while a private data copy
raises it to 8--12\,s. Sub-second reads rarely break even; long tests clear
either bound.

\begin{table}[t]
\centering
\caption{Microbenchmark of runtime primitive costs.}
\label{tab:primitive-costs}
\small
\begin{tabularx}{\columnwidth}{@{}X r@{}}
\toprule
Primitive & Elapsed time \\
\midrule
Read-only E2B overlay fork & 0.08\,s \\
Fork with private data copy & 3--5.5\,s \\
Full snapshot creation & 10--60\,s \\
Warm reconnect & 0.5--3\,s \\
Trace IR recording & 0.1--1\,s \\
Canonical verification & $<50$\,ms \\
Dependency checking & 50\,ms--2\,s \\
Commit, promotion, or recovery & 0.5--5\,s \\
Journal flush & 1--50\,ms \\
\bottomrule
\end{tabularx}
\end{table}

End-to-end overhead divides into drafter calls (28\%), sandbox management
(32\%), validation and tracing (15\%), and recovery (25\%). These costs buy
single-session latency rather than throughput-normalized efficiency: the
additional speculation GPU could instead serve more Serial Pi sessions. The
main agent consumes nearly the same prompt-plus-completion tokens as Serial Pi
(8.3M versus 8.2M per task), while speculation adds 1.6M drafter tokens, 1{,}480
drafter GPU-seconds, and raises sandbox CPU time from 2{,}760 to 4{,}310 seconds
per task. As a coarse proxy, \system spends about 2.1 seconds of speculative
compute for each wall-clock second removed.
Concretely, drafter inference adds 1{,}480 GPU-seconds per task against the
main agent's 1{,}640, so the speculation GPU budget is 0.9$\times$ one
Serial~Pi session. In a throughput-optimal deployment the drafter GPU budget
could serve roughly one additional session instead; \system targets the
complementary single-session interactive regime where the user waits for one
task result.

The main agent performs identical decoding work under both configurations
(Appendix~\ref{app:eval-parity}), so the removed wall clock comes from tool
executions. A misprediction costs one overlay; recovery discards it in
0.5--5\,s regardless of chain length, so the cost tracks the isolation
mechanism rather than misprediction depth.

Appendix~\ref{app:eval-extra} ablates the two-class operand model: both
single-version alternatives preserve zero false accepts but raise normalized
latency from 0.58 to 0.68 (permissive) and 0.85 (conservative), confirming
that commit validation supplies safety while the two-class split supplies
profitable reuse.

\begin{figure}[t]
\centering
\evalplot{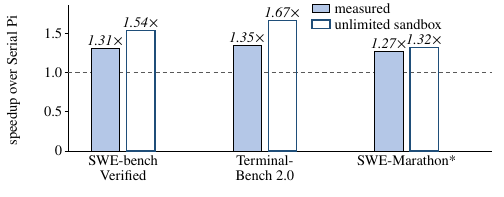}
\caption{Measured speedup and the unlimited-sandbox dependence roofline over
Serial Pi. }
\Description{A bar chart with three groups, SWE-bench Verified, Terminal-Bench 2.0 and starred SWE-Marathon, each with a filled bar for measured speedup and a hollow bar for the unlimited-sandbox roofline; values 1.31 and 1.54, 1.35 and 1.67, and 1.27 and 1.32.}
\label{fig:eval-limit}
\end{figure}

Figure~\ref{fig:eval-limit} compares measured speedup with a single
unlimited-sandbox dependence roofline. The roofline keeps the same main-agent
decoding path and recorded dependence structure, then removes the sandbox-slot
cap. Under that bound, SWE-bench Verified, Terminal-Bench~2.0, and
SWE-Marathon reach 1.54$\times$, 1.67$\times$, and 1.32$\times$,
respectively. The measured bars reach 85\%, 81\%, and 96\% of those bounds,
so the remaining gap is an allocation and runtime-overhead gap rather than a
separate response-reuse ceiling. The depth/width sweep in
Appendix~\ref{app:eval-extra} reports the same allocation effect on the
SWE-Marathon trace: depth saturates at $K{=}4$, beyond which no additional candidate
commits (consistent with the chain-depth distribution of
Figure~\ref{fig:eval-mechanisms}), and width saturates at sixteen
slots---twice what our E2B account can actually run---because the
producer chain must still execute in order.

In summary, \system trades 0.9$\times$ of one Serial~Pi session's GPU budget
and 1.56$\times$ sandbox CPU time for a 1.27--1.35$\times$ single-session
latency reduction, capturing 81--96\% of the dependence-limited roofline.
Overhead is dominated by sandbox management (32\%) and recovery (25\%), both
set by infrastructure constants. The primary path to further gains is drafter
accuracy: each ten-point increase in acceptance adds 0.047 of
normalized-latency reduction with no sign of saturation.

\section{Related Work}
\label{sec:related}

\inlinesection{Out-of-order and speculative execution.}
Tomasulo's algorithm~\cite{tomasulo1967efficient} and the reorder buffer~\cite{smith1988precise} enable hardware to execute instructions out of program order while committing results in order.
Thread-level speculation~\cite{sohi1995multiscalar,steffan2000tls} extends this model to coarser thread granularity, deriving read/write sets from hardware snoops.
\system transplants the same execute-out-of-order, commit-in-order shape to agent tool calls, replacing register/cache tracking with syscall-level file-version evidence and hardware rollback with COW-sandbox discard.

\inlinesection{Speculative drafting in agent systems.}
Speculative decoding~\cite{leviathan2023fast,chen2023speculative} drafts tokens with a smaller model; \system lifts the idea to the \emph{action} level.
SPORK~\cite{bai2026spork}, DualSpec~\cite{zhong2026dualspec}, AOSpec~\cite{chen2026aospec}, and Speculative Actions~\cite{ye2026speculativeactions} treat the tool call as the atomic unit of speculation: a mispredicted call discards all downstream work.
\system differs in tracking per-file version evidence so that only calls whose operands actually changed require re-execution (Section~\ref{sec:exp_q1} compares against PASTE's prefetch-only baseline~\cite{sui2026paste}).

\inlinesection{Transactional and parallel agent execution.}
Atomix~\cite{mohammadi2026atomix}, SagaLLM~\cite{chang2025sagallm}, and agentic-transaction proposals~\cite{agentictransaction2026} apply database-style transactions to agent tool calls for reliability.
\system addresses a different axis---latency rather than fault tolerance---using optimistic concurrency control~\cite{kung1981optimistic,herlihy1993transactional} over a transaction body discovered incrementally.
hS~\cite{hs2026} and LLMCompiler~\cite{kim2024llmcompiler} parallelize calls when dependencies are known statically; \system dispatches calls whose independence is \emph{unknown} and resolves correctness at commit through runtime evidence.

\section{Conclusion}
\label{sec:conclusion}

\system applies speculative out-of-order issue with in-order commit to agent tool execution, turning cross-turn observation stalls into overlap that same-turn parallel dispatch cannot reach. COW overlays, Trace~IR tracking, and validated-prefix commit ensure that no speculative result becomes task-visible without justification from committed state. On SWE-bench Verified, Terminal-Bench~2.0, and SWE-Marathon, \system achieves 1.27--1.35$\times$ speedups with zero false accepts across 4{,}010 audited records. As tool latency increasingly dominates agent wall time, the boundary between speculative prediction and validated commit will become a key design axis for agent runtimes.

\bibliographystyle{ACM-Reference-Format}
\bibliography{refs}


\begin{thebibliography}{33}


\ifx \showCODEN    \undefined \def \showCODEN     #1{\unskip}     \fi
\ifx \showISBNx    \undefined \def \showISBNx     #1{\unskip}     \fi
\ifx \showISBNxiii \undefined \def \showISBNxiii  #1{\unskip}     \fi
\ifx \showISSN     \undefined \def \showISSN      #1{\unskip}     \fi
\ifx \showLCCN     \undefined \def \showLCCN      #1{\unskip}     \fi
\ifx \shownote     \undefined \def \shownote      #1{#1}          \fi
\ifx \showarticletitle \undefined \def \showarticletitle #1{#1}   \fi
\ifx \showURL      \undefined \def \showURL       {\relax}        \fi
\providecommand\bibfield[2]{#2}
\providecommand\bibinfo[2]{#2}
\providecommand\natexlab[1]{#1}
\providecommand\showeprint[2][]{arXiv:#2}

\bibitem[Abts et~al\mbox{.}(2022)]%
        {abts2022groq}
\bibfield{author}{\bibinfo{person}{Dennis Abts}, \bibinfo{person}{Garrin
  Kimmell}, \bibinfo{person}{Andrew Ling}, \bibinfo{person}{John Kim},
  \bibinfo{person}{Matt Boyd}, \bibinfo{person}{Andrew Bitar},
  \bibinfo{person}{Sahil Parmar}, \bibinfo{person}{Ibrahim Ahmed},
  \bibinfo{person}{Roberto DiCecco}, \bibinfo{person}{David Han},
  \bibinfo{person}{John Thompson}, \bibinfo{person}{Michael Bye},
  \bibinfo{person}{Jennifer Hwang}, \bibinfo{person}{Jeremy Fowers},
  \bibinfo{person}{Peter Lillian}, \bibinfo{person}{Ashwin Murthy},
  \bibinfo{person}{Elyas Mehtabuddin}, \bibinfo{person}{Chetan Tekur},
  \bibinfo{person}{Thomas Sohmers}, \bibinfo{person}{Kris Kang},
  \bibinfo{person}{Stephen Maresh}, {and} \bibinfo{person}{Jonathan Ross}.}
  \bibinfo{year}{2022}\natexlab{}.
\newblock \showarticletitle{A Software-Defined Tensor Streaming Multiprocessor
  for Large-Scale Machine Learning}. In \bibinfo{booktitle}{\emph{Proceedings
  of the 49th Annual International Symposium on Computer Architecture (ISCA)}}.
  \bibinfo{pages}{567--580}.
\newblock


\bibitem[Bai et~al\mbox{.}(2026)]%
        {bai2026spork}
\bibfield{author}{\bibinfo{person}{Huajun Bai}, \bibinfo{person}{Weiwei Lv},
  \bibinfo{person}{Huichuan Zheng}, \bibinfo{person}{Youyou Lu}, {and}
  \bibinfo{person}{Jiwu Shu}.} \bibinfo{year}{2026}\natexlab{}.
\newblock \showarticletitle{{SPORK}: Self-Speculative Forking to Accelerate
  Agentic {LLM} Inference}.
\newblock \bibinfo{journal}{\emph{arXiv preprint arXiv:2607.03333}}
  (\bibinfo{year}{2026}).
\newblock
\urldef\tempurl%
\url{https://arxiv.org/abs/2607.03333}
\showURL{%
\tempurl}


\bibitem[{Cerebras Systems}(2026)]%
        {cerebras2026inference}
\bibfield{author}{\bibinfo{person}{{Cerebras Systems}}.}
  \bibinfo{year}{2026}\natexlab{}.
\newblock \bibinfo{title}{Cerebras Inference: Frontier Models at Over 2,000
  Tokens per Second on Wafer-Scale Hardware}.
\newblock \bibinfo{howpublished}{\url{https://www.cerebras.ai/inference}}.
\newblock
\newblock
\shownote{Accessed September 2026}.


\bibitem[Chang and Geng(2025)]%
        {chang2025sagallm}
\bibfield{author}{\bibinfo{person}{Edward~Y. Chang} {and}
  \bibinfo{person}{Longling Geng}.} \bibinfo{year}{2025}\natexlab{}.
\newblock \showarticletitle{{SagaLLM}: Context Management, Validation, and
  Transaction Guarantees for Multi-Agent {LLM} Planning}.
\newblock \bibinfo{journal}{\emph{Proceedings of the VLDB Endowment}}
  \bibinfo{volume}{18}, \bibinfo{number}{12} (\bibinfo{year}{2025}),
  \bibinfo{pages}{4874--4886}.
\newblock
\href{https://doi.org/10.14778/3750601.3750611}{doi:\nolinkurl{10.14778/3750601.3750611}}


\bibitem[Chen et~al\mbox{.}(2023)]%
        {chen2023speculative}
\bibfield{author}{\bibinfo{person}{Charlie Chen}, \bibinfo{person}{Sebastian
  Borgeaud}, \bibinfo{person}{Geoffrey Irving}, \bibinfo{person}{Jean-Baptiste
  Lespiau}, \bibinfo{person}{Laurent Sifre}, {and} \bibinfo{person}{John
  Jumper}.} \bibinfo{year}{2023}\natexlab{}.
\newblock \showarticletitle{Accelerating Large Language Model Decoding with
  Speculative Sampling}.
\newblock \bibinfo{journal}{\emph{arXiv preprint arXiv:2302.01318}}
  (\bibinfo{year}{2023}).
\newblock
\urldef\tempurl%
\url{https://arxiv.org/abs/2302.01318}
\showURL{%
\tempurl}


\bibitem[Chen et~al\mbox{.}(2026)]%
        {chen2026aospec}
\bibfield{author}{\bibinfo{person}{Hao~Mark Chen}, \bibinfo{person}{Jinnan
  Guo}, \bibinfo{person}{Wayne Luk}, {and} \bibinfo{person}{Hongxiang Fan}.}
  \bibinfo{year}{2026}\natexlab{}.
\newblock \showarticletitle{{AOSpec}: Action and Observation Co-Speculation for
  Low-Latency Agent Serving}.
\newblock \bibinfo{journal}{\emph{arXiv preprint arXiv:2608.00881}}
  (\bibinfo{year}{2026}).
\newblock
\urldef\tempurl%
\url{https://arxiv.org/abs/2608.00881}
\showURL{%
\tempurl}


\bibitem[Curtsinger and Barowy(2022)]%
        {curtsinger2022riker}
\bibfield{author}{\bibinfo{person}{Charlie Curtsinger} {and}
  \bibinfo{person}{Daniel~W. Barowy}.} \bibinfo{year}{2022}\natexlab{}.
\newblock \showarticletitle{Riker: Always-Correct and Fast Incremental Builds
  from Simple Specifications}. In \bibinfo{booktitle}{\emph{Proceedings of the
  2022 USENIX Annual Technical Conference}}. \bibinfo{publisher}{USENIX
  Association}, \bibinfo{pages}{885--898}.
\newblock
\urldef\tempurl%
\url{https://www.usenix.org/conference/atc22/presentation/curtsinger}
\showURL{%
\tempurl}


\bibitem[Desai et~al\mbox{.}(2026)]%
        {swemarathon2026}
\bibfield{author}{\bibinfo{person}{Rishi Desai}, \bibinfo{person}{Jesse Hu},
  \bibinfo{person}{Joan Cabezas}, \bibinfo{person}{Neel Harsola},
  \bibinfo{person}{Pratyush Shukla}, \bibinfo{person}{Roey~Ben Chaim},
  \bibinfo{person}{Adnan~El Assadi}, \bibinfo{person}{Omkaar~Mukund Kamath},
  \bibinfo{person}{Fenil Faldu}, \bibinfo{person}{Prannay Hebbar},
  \bibinfo{person}{Jiankai Sun}, \bibinfo{person}{Yiyuan Li},
  \bibinfo{person}{Pramod Srinivasan}, \bibinfo{person}{Ishan Gupta},
  \bibinfo{person}{Christopher Settles}, \bibinfo{person}{Daniel Wang},
  \bibinfo{person}{Derek Chen}, \bibinfo{person}{Pranav Raja},
  \bibinfo{person}{Albert Liu}, \bibinfo{person}{Marek {\v S}uppa},
  \bibinfo{person}{Nevasini Sasikumar}, \bibinfo{person}{Luyang Kong},
  \bibinfo{person}{Erik Quintanilla}, \bibinfo{person}{Xiangyi Li},
  \bibinfo{person}{Ivan Bercovich}, {and} \bibinfo{person}{Steven Dillmann}.}
  \bibinfo{year}{2026}\natexlab{}.
\newblock \bibinfo{title}{{SWE-Marathon}: Can Agents Autonomously Complete
  Ultra-Long-Horizon Software Work?}
\newblock \bibinfo{howpublished}{\url{https://arxiv.org/abs/2606.07682}}.
\newblock
\urldef\tempurl%
\url{https://arxiv.org/abs/2606.07682}
\showURL{%
\tempurl}
\newblock
\shownote{Benchmark and evaluation code available at
  \url{https://github.com/abundant-ai/swe-marathon}}.


\bibitem[{E2B}(2026)]%
        {e2b2026snapshots}
\bibfield{author}{\bibinfo{person}{{E2B}}.} \bibinfo{year}{2026}\natexlab{}.
\newblock \bibinfo{title}{Sandbox Snapshots}.
\newblock \bibinfo{howpublished}{\url{https://docs.e2b.dev/sandbox/snapshots}}.
\newblock
\newblock
\shownote{Accessed September 2026}.


\bibitem[{Earendil Works}(2025)]%
        {earendilpi2025}
\bibfield{author}{\bibinfo{person}{{Earendil Works}}.}
  \bibinfo{year}{2025}\natexlab{}.
\newblock \bibinfo{title}{{Pi}: An Open-Source Coding Agent Harness}.
\newblock \bibinfo{howpublished}{\url{https://github.com/earendil-works/pi}}.
\newblock
\newblock
\shownote{Open-source TypeScript monorepo; npm package
  \texttt{@earendil-works/pi-coding-agent}}.


\bibitem[Herlihy and Moss(1993)]%
        {herlihy1993transactional}
\bibfield{author}{\bibinfo{person}{Maurice Herlihy} {and}
  \bibinfo{person}{J.~Eliot~B. Moss}.} \bibinfo{year}{1993}\natexlab{}.
\newblock \showarticletitle{Transactional Memory: Architectural Support for
  Lock-Free Data Structures}. In \bibinfo{booktitle}{\emph{Proceedings of the
  20th Annual International Symposium on Computer Architecture (ISCA)}}.
  \bibinfo{pages}{289--300}.
\newblock


\bibitem[Ji et~al\mbox{.}(2026)]%
        {ji2026speculate}
\bibfield{author}{\bibinfo{person}{Jiabao Ji}, \bibinfo{person}{Yujian Liu},
  \bibinfo{person}{Li An}, \bibinfo{person}{Rohit Jain},
  \bibinfo{person}{Gungor Polatkan}, \bibinfo{person}{Siyu Zhu}, {and}
  \bibinfo{person}{Shiyu Chang}.} \bibinfo{year}{2026}\natexlab{}.
\newblock \showarticletitle{Speculate While You Reason: Teaching Agents to
  Predict Their Next Tool Call via Joint Agent-Speculator {RL}}.
\newblock \bibinfo{journal}{\emph{arXiv preprint arXiv:2607.25816}}
  (\bibinfo{year}{2026}).
\newblock
\urldef\tempurl%
\url{https://arxiv.org/abs/2607.25816}
\showURL{%
\tempurl}


\bibitem[Jimenez et~al\mbox{.}(2024)]%
        {jimenez2024swebench}
\bibfield{author}{\bibinfo{person}{Carlos~E. Jimenez}, \bibinfo{person}{John
  Yang}, \bibinfo{person}{Alexander Wettig}, \bibinfo{person}{Shunyu Yao},
  \bibinfo{person}{Kexin Pei}, \bibinfo{person}{Ofir Press}, {and}
  \bibinfo{person}{Karthik Narasimhan}.} \bibinfo{year}{2024}\natexlab{}.
\newblock \showarticletitle{{SWE-bench}: Can Language Models Resolve Real-World
  {GitHub} Issues?}
\newblock \bibinfo{journal}{\emph{arXiv preprint arXiv:2310.06770}}
  (\bibinfo{year}{2024}).
\newblock
\urldef\tempurl%
\url{https://arxiv.org/abs/2310.06770}
\showURL{%
\tempurl}


\bibitem[Kim et~al\mbox{.}(2024)]%
        {kim2024llmcompiler}
\bibfield{author}{\bibinfo{person}{Sehoon Kim}, \bibinfo{person}{Suhong Moon},
  \bibinfo{person}{Ryan Tabrizi}, \bibinfo{person}{Nicholas Lee},
  \bibinfo{person}{Michael~W. Mahoney}, \bibinfo{person}{Kurt Keutzer}, {and}
  \bibinfo{person}{Amir Gholami}.} \bibinfo{year}{2024}\natexlab{}.
\newblock \showarticletitle{An {LLM} Compiler for Parallel Function Calling}.
  In \bibinfo{booktitle}{\emph{Proceedings of the 41st International Conference
  on Machine Learning}}.
\newblock
\urldef\tempurl%
\url{https://arxiv.org/abs/2312.04511}
\showURL{%
\tempurl}


\bibitem[Kung and Robinson(1981)]%
        {kung1981optimistic}
\bibfield{author}{\bibinfo{person}{H.~T. Kung} {and} \bibinfo{person}{John~T.
  Robinson}.} \bibinfo{year}{1981}\natexlab{}.
\newblock \showarticletitle{On Optimistic Methods for Concurrency Control}.
\newblock \bibinfo{journal}{\emph{ACM Transactions on Database Systems}}
  \bibinfo{volume}{6}, \bibinfo{number}{2} (\bibinfo{year}{1981}),
  \bibinfo{pages}{213--226}.
\newblock


\bibitem[Leviathan et~al\mbox{.}(2023)]%
        {leviathan2023fast}
\bibfield{author}{\bibinfo{person}{Yaniv Leviathan}, \bibinfo{person}{Matan
  Kalman}, {and} \bibinfo{person}{Yossi Matias}.}
  \bibinfo{year}{2023}\natexlab{}.
\newblock \showarticletitle{Fast Inference from Transformers via Speculative
  Decoding}. In \bibinfo{booktitle}{\emph{Proceedings of the 40th International
  Conference on Machine Learning}} \emph{(\bibinfo{series}{Proceedings of
  Machine Learning Research}, Vol.~\bibinfo{volume}{202})}.
  \bibinfo{pages}{19274--19286}.
\newblock
\urldef\tempurl%
\url{https://proceedings.mlr.press/v202/leviathan23a.html}
\showURL{%
\tempurl}


\bibitem[Liargkovas et~al\mbox{.}(2026)]%
        {hs2026}
\bibfield{author}{\bibinfo{person}{Georgios Liargkovas}, \bibinfo{person}{Di
  Jin}, \bibinfo{person}{Tianyu~(Ezri) Zhu}, \bibinfo{person}{Dan Liu},
  \bibinfo{person}{A.~Bolun Thompson}, \bibinfo{person}{Anirudh Narsipur},
  \bibinfo{person}{Seong-Heon Jung}, \bibinfo{person}{Siddhartha Prasad},
  \bibinfo{person}{Diomidis Spinellis}, \bibinfo{person}{Michael Greenberg},
  \bibinfo{person}{Konstantinos Kallas}, {and} \bibinfo{person}{Nikos
  Vasilakis}.} \bibinfo{year}{2026}\natexlab{}.
\newblock \showarticletitle{{hS}: Speculative Script Reordering at Subprocess
  Granularity}. In \bibinfo{booktitle}{\emph{20th USENIX Symposium on Operating
  Systems Design and Implementation}}. \bibinfo{publisher}{USENIX Association},
  \bibinfo{address}{Seattle, WA}, \bibinfo{pages}{665--681}.
\newblock
\urldef\tempurl%
\url{https://www.usenix.org/conference/osdi26/presentation/liargkovas}
\showURL{%
\tempurl}


\bibitem[Lipasti and Shen(1996)]%
        {lipasti1996exceeding}
\bibfield{author}{\bibinfo{person}{Mikko~H. Lipasti} {and}
  \bibinfo{person}{John~Paul Shen}.} \bibinfo{year}{1996}\natexlab{}.
\newblock \showarticletitle{Exceeding the Dataflow Limit via Value Prediction}.
  In \bibinfo{booktitle}{\emph{Proceedings of the 29th Annual IEEE/ACM
  International Symposium on Microarchitecture}}. \bibinfo{pages}{226--237}.
\newblock
\href{https://doi.org/10.1109/MICRO.1996.566464}{doi:\nolinkurl{10.1109/MICRO.1996.566464}}


\bibitem[Mialon et~al\mbox{.}(2023)]%
        {mialon2023gaia}
\bibfield{author}{\bibinfo{person}{Gr{\'e}goire Mialon},
  \bibinfo{person}{Cl{\'e}mentine Fourrier}, \bibinfo{person}{Craig Swift},
  \bibinfo{person}{Thomas Wolf}, \bibinfo{person}{Yann LeCun}, {and}
  \bibinfo{person}{Thomas Scialom}.} \bibinfo{year}{2023}\natexlab{}.
\newblock \showarticletitle{{GAIA}: A Benchmark for General {AI} Assistants}.
\newblock \bibinfo{journal}{\emph{arXiv preprint arXiv:2311.12983}}
  (\bibinfo{year}{2023}).
\newblock
\urldef\tempurl%
\url{https://arxiv.org/abs/2311.12983}
\showURL{%
\tempurl}


\bibitem[Mohammadi et~al\mbox{.}(2026)]%
        {mohammadi2026atomix}
\bibfield{author}{\bibinfo{person}{Bardia Mohammadi}, \bibinfo{person}{Nearchos
  Potamitis}, \bibinfo{person}{Lars Klein}, \bibinfo{person}{Akhil Arora},
  {and} \bibinfo{person}{Laurent Bindschaedler}.}
  \bibinfo{year}{2026}\natexlab{}.
\newblock \showarticletitle{Atomix: Timely, Transactional Tool Use for Reliable
  Agentic Workflows}.
\newblock \bibinfo{journal}{\emph{arXiv preprint arXiv:2602.14849}}
  (\bibinfo{year}{2026}).
\newblock
\urldef\tempurl%
\url{https://arxiv.org/abs/2602.14849}
\showURL{%
\tempurl}


\bibitem[{OpenAI}(2024)]%
        {openai2024swebenchverified}
\bibfield{author}{\bibinfo{person}{{OpenAI}}.} \bibinfo{year}{2024}\natexlab{}.
\newblock \bibinfo{title}{Introducing {SWE-bench Verified}}.
\newblock
  \bibinfo{howpublished}{\url{https://openai.com/index/introducing-swe-bench-verified/}}.
\newblock


\bibitem[Schick et~al\mbox{.}(2023)]%
        {schick2023toolformer}
\bibfield{author}{\bibinfo{person}{Timo Schick}, \bibinfo{person}{Jane
  Dwivedi-Yu}, \bibinfo{person}{Roberto Dess{\`i}}, \bibinfo{person}{Roberta
  Raileanu}, \bibinfo{person}{Maria Lomeli}, \bibinfo{person}{Eric Hambro},
  \bibinfo{person}{Luke Zettlemoyer}, \bibinfo{person}{Nicola Cancedda}, {and}
  \bibinfo{person}{Thomas Scialom}.} \bibinfo{year}{2023}\natexlab{}.
\newblock \showarticletitle{Toolformer: Language Models Can Teach Themselves to
  Use Tools}. In \bibinfo{booktitle}{\emph{Advances in Neural Information
  Processing Systems}}.
\newblock
\urldef\tempurl%
\url{https://arxiv.org/abs/2302.04761}
\showURL{%
\tempurl}


\bibitem[Smith and Pleszkun(1988)]%
        {smith1988precise}
\bibfield{author}{\bibinfo{person}{James~E. Smith} {and}
  \bibinfo{person}{Andrew~R. Pleszkun}.} \bibinfo{year}{1988}\natexlab{}.
\newblock \showarticletitle{Implementing Precise Interrupts in Pipelined
  Processors}.
\newblock \bibinfo{journal}{\emph{IEEE Trans. Comput.}} \bibinfo{volume}{37},
  \bibinfo{number}{5} (\bibinfo{year}{1988}), \bibinfo{pages}{562--573}.
\newblock
\href{https://doi.org/10.1109/12.4607}{doi:\nolinkurl{10.1109/12.4607}}


\bibitem[Sohi et~al\mbox{.}(1995)]%
        {sohi1995multiscalar}
\bibfield{author}{\bibinfo{person}{Gurindar~S. Sohi}, \bibinfo{person}{Scott~E.
  Breach}, {and} \bibinfo{person}{T.~N. Vijaykumar}.}
  \bibinfo{year}{1995}\natexlab{}.
\newblock \showarticletitle{Multiscalar Processors}. In
  \bibinfo{booktitle}{\emph{Proceedings of the 22nd Annual International
  Symposium on Computer Architecture (ISCA)}}. \bibinfo{pages}{414--425}.
\newblock


\bibitem[Steffan et~al\mbox{.}(2000)]%
        {steffan2000tls}
\bibfield{author}{\bibinfo{person}{J.~Gregory Steffan},
  \bibinfo{person}{Christopher~B. Colohan}, \bibinfo{person}{Antonia Zhai},
  {and} \bibinfo{person}{Todd~C. Mowry}.} \bibinfo{year}{2000}\natexlab{}.
\newblock \showarticletitle{A Scalable Approach to Thread-Level Speculation}.
  In \bibinfo{booktitle}{\emph{Proceedings of the 27th Annual International
  Symposium on Computer Architecture (ISCA)}}. \bibinfo{pages}{1--12}.
\newblock


\bibitem[Sui et~al\mbox{.}(2026)]%
        {sui2026paste}
\bibfield{author}{\bibinfo{person}{Yifan Sui}, \bibinfo{person}{Han Zhao},
  \bibinfo{person}{Rui Ma}, \bibinfo{person}{Zhiyuan He}, \bibinfo{person}{Hao
  Wang}, \bibinfo{person}{Jianxun Li}, \bibinfo{person}{Kaiqiang Xu},
  \bibinfo{person}{Kai Chen}, {and} \bibinfo{person}{Yuqing Yang}.}
  \bibinfo{year}{2026}\natexlab{}.
\newblock \showarticletitle{Parallelizing Tool Execution and {LLM} Generation
  for Low-Latency Agent Serving}.
\newblock \bibinfo{journal}{\emph{arXiv preprint arXiv:2603.18897}}
  (\bibinfo{year}{2026}).
\newblock
\urldef\tempurl%
\url{https://arxiv.org/abs/2603.18897}
\showURL{%
\tempurl}


\bibitem[Sun et~al\mbox{.}(2026)]%
        {agentictransaction2026}
\bibfield{author}{\bibinfo{person}{Zhaoyan Sun}, \bibinfo{person}{Xiaoxiao
  Wang}, {and} \bibinfo{person}{Guoliang Li}.} \bibinfo{year}{2026}\natexlab{}.
\newblock \showarticletitle{Agentic Transaction: Towards {ACID}-Compliant Agent
  Systems}.
\newblock \bibinfo{journal}{\emph{arXiv preprint arXiv:2608.13900}}
  (\bibinfo{year}{2026}).
\newblock
\urldef\tempurl%
\url{https://arxiv.org/abs/2608.13900}
\showURL{%
\tempurl}


\bibitem[{The Terminal-Bench Team}(2025)]%
        {tb2}
\bibfield{author}{\bibinfo{person}{{The Terminal-Bench Team}}.}
  \bibinfo{year}{2025}\natexlab{}.
\newblock \bibinfo{title}{{Terminal-Bench 2.0}}.
\newblock
  \bibinfo{howpublished}{\url{https://github.com/harbor-framework/terminal-bench-2}}.
\newblock
\newblock
\shownote{Laude Institute and Harbor Framework}.


\bibitem[Tomasulo(1967)]%
        {tomasulo1967efficient}
\bibfield{author}{\bibinfo{person}{Robert~M. Tomasulo}.}
  \bibinfo{year}{1967}\natexlab{}.
\newblock \showarticletitle{An Efficient Algorithm for Exploiting Multiple
  Arithmetic Units}.
\newblock \bibinfo{journal}{\emph{IBM Journal of Research and Development}}
  \bibinfo{volume}{11}, \bibinfo{number}{1} (\bibinfo{year}{1967}),
  \bibinfo{pages}{25--33}.
\newblock
\href{https://doi.org/10.1147/rd.111.0025}{doi:\nolinkurl{10.1147/rd.111.0025}}


\bibitem[Yao et~al\mbox{.}(2023)]%
        {yao2023react}
\bibfield{author}{\bibinfo{person}{Shunyu Yao}, \bibinfo{person}{Jeffrey Zhao},
  \bibinfo{person}{Dian Yu}, \bibinfo{person}{Nan Du}, \bibinfo{person}{Izhak
  Shafran}, \bibinfo{person}{Karthik Narasimhan}, {and} \bibinfo{person}{Yuan
  Cao}.} \bibinfo{year}{2023}\natexlab{}.
\newblock \showarticletitle{{ReAct}: Synergizing Reasoning and Acting in
  Language Models}. In \bibinfo{booktitle}{\emph{International Conference on
  Learning Representations}}.
\newblock
\urldef\tempurl%
\url{https://openreview.net/forum?id=WE_vluYUL-X}
\showURL{%
\tempurl}


\bibitem[Ye et~al\mbox{.}(2026)]%
        {ye2026speculativeactions}
\bibfield{author}{\bibinfo{person}{Naimeng Ye}, \bibinfo{person}{Arnav Ahuja},
  \bibinfo{person}{Georgios Liargkovas}, \bibinfo{person}{Yunan Lu},
  \bibinfo{person}{Kostis Kaffes}, {and} \bibinfo{person}{Tianyi Peng}.}
  \bibinfo{year}{2026}\natexlab{}.
\newblock \showarticletitle{Speculative Actions: A Lossless Framework for
  Faster Agentic Systems}.
\newblock \bibinfo{journal}{\emph{arXiv preprint arXiv:2510.04371}}
  (\bibinfo{year}{2026}).
\newblock
\urldef\tempurl%
\url{https://arxiv.org/abs/2510.04371}
\showURL{%
\tempurl}


\bibitem[Zhong et~al\mbox{.}(2026)]%
        {zhong2026dualspec}
\bibfield{author}{\bibinfo{person}{Shuzhang Zhong}, \bibinfo{person}{Baotong
  Lu}, \bibinfo{person}{Qi Chen}, \bibinfo{person}{Chuanjie Liu},
  \bibinfo{person}{Fan Yang}, {and} \bibinfo{person}{Meng Li}.}
  \bibinfo{year}{2026}\natexlab{}.
\newblock \showarticletitle{{DualSpec}: Accelerating Deep Research Agents via
  Dual-Process Action Speculation}.
\newblock \bibinfo{journal}{\emph{arXiv preprint arXiv:2603.07416}}
  (\bibinfo{year}{2026}).
\newblock
\urldef\tempurl%
\url{https://arxiv.org/abs/2603.07416}
\showURL{%
\tempurl}


\bibitem[Zuo et~al\mbox{.}(2026)]%
        {zuo2026qwenagentworld}
\bibfield{author}{\bibinfo{person}{Yuxin Zuo}, \bibinfo{person}{Zikai Xiao},
  \bibinfo{person}{Li Sheng}, \bibinfo{person}{Fei Huang},
  \bibinfo{person}{Jianhong Tu}, \bibinfo{person}{Yuxuan Liu},
  \bibinfo{person}{Tianyi Tang}, \bibinfo{person}{Xiaomeng Hu},
  \bibinfo{person}{Yang Su}, \bibinfo{person}{Qingfeng Lan},
  \bibinfo{person}{Yantao Liu}, \bibinfo{person}{Qin Zhu},
  \bibinfo{person}{Yinger Zhang}, \bibinfo{person}{Bowen Yu},
  \bibinfo{person}{Haiquan Zhao}, \bibinfo{person}{Haiyang Xu},
  \bibinfo{person}{Jianxin Yang}, \bibinfo{person}{Jiayang Cheng},
  \bibinfo{person}{Junyang Wang}, \bibinfo{person}{Lianghao Deng},
  \bibinfo{person}{Mingfeng Xue}, \bibinfo{person}{Tianyi Bai},
  \bibinfo{person}{Yang Fan}, \bibinfo{person}{Yubo Ma},
  \bibinfo{person}{Yucheng Li}, \bibinfo{person}{Zeyu Cui},
  \bibinfo{person}{Zhihai Wang}, \bibinfo{person}{Zhihui Xie},
  \bibinfo{person}{Zhuorui Ye}, \bibinfo{person}{An Yang},
  \bibinfo{person}{Dayiheng Liu}, \bibinfo{person}{Jingren Zhou}, {and}
  \bibinfo{person}{Ning Ding}.} \bibinfo{year}{2026}\natexlab{}.
\newblock \showarticletitle{{Qwen-AgentWorld}: Language World Models for
  General Agents}.
\newblock \bibinfo{journal}{\emph{arXiv preprint arXiv:2606.24597}}
  (\bibinfo{year}{2026}).
\newblock
\urldef\tempurl%
\url{https://arxiv.org/abs/2606.24597}
\showURL{%
\tempurl}


\end{thebibliography}

\clearpage
\appendix
\section{Tool Observation and Replay Details}
\label{app:impl-details}

\subsection{Tool Wrapper API}
\label{app:wrapper-schema}

Tool wrappers implement the correctness-preserving commit rule (Section~\ref{sec:commit-correctness}) through a typed API over ATIF records (Section~\ref{sec:implementation}): classify the call's safety class, capture an input snapshot, execute in an overlay, canonicalize observations, emit Trace IR records, run commit validation, and promote or replay validated effects. The test class canonicalization, for example, retains \texttt{exit\_code}, \texttt{failed\_tests}, \texttt{error\_types}, source locations, and artifact hashes. It drops \texttt{duration}, process identifiers, and absolute temporary paths. \texttt{schema\_version} pins the canonicalization rules so a future schema change forces a miss rather than a silent mismatch. Fields not declared as droppable must match exactly after canonicalization; wrapper parse failure, schema-version mismatch, or untraceable effects force a miss or speculation barrier.

Table~\ref{tab:tool-wrapper-classes} summarizes the implemented wrapper surface. The prototype implements no remote-endpoint wrapper, so unversioned and non-idempotent network calls are speculation barriers.

\begin{table}[t]
\centering
\caption{Implemented tool-wrapper classes and comparators. Test, compile, search, and general \texttt{bash} commands share the \texttt{bash} wrapper; class-specific projected fields and comparators are implemented through the canonicalization policy (Section~\ref{sec:commit-correctness}).}
\label{tab:tool-wrapper-classes}
\scriptsize
\begin{tabularx}{\columnwidth}{@{}l X l@{}}
\toprule
Tool class & Projected fields & Comparator \\
\midrule
\texttt{read} & path, existence, content hash, path-absence dependencies, mtime policy & Exact digest \\
\texttt{edit}/\texttt{write} & target paths, patch/effect digest, written bytes, post-state digest & Exact digest \\
test/compile & exit code, failed tests, error types, source locations, artifact hashes & Field-wise exact \\
search & query, matched paths, line numbers, content hashes, stable ordering policy & Canonical set \\
\texttt{bash} & exit code, declared stdout/stderr fields, effects, opaque flag & Tool-specific \\
opaque/API & none; optional prepare intent only & Speculation barrier \\
\bottomrule
\end{tabularx}
\end{table}

\subsection{ATIF Records}
\label{app:atif}

Both adapter operations (Section~\ref{sec:impl-harness}) normalize harness-native records into ATIF, a small action/observation interchange format. The dispatch side runs after Pi parses a model-produced action and before the tool executes; the observation side runs after the tool returns and before Pi appends the result to the model context, and it decides whether dependent speculative work stays eligible. Pi continues to build prompts, serialize tool schemas, call the model provider, and parse tool calls through its own interfaces. An ATIF action contains the tool name, canonical JSON arguments, working directory, and stable environment; an ATIF observation contains the model-visible result, observation schema version, and timing metadata. The Run-Ahead Controller tracks overlay lineage ($L_i$ in the Trace IR record of Appendix~\ref{app:trace-ir-schema}) through the candidate tree separately from the action record. The same converter normalizes the 212 offline reference trajectories of Section~\ref{sec:introduction}. The adapter also records the harness event that caused each transition, which lets the runtime distinguish a candidate dispatch from a committed observation. All downstream components consume ATIF rather than Pi-native structures, including the drafters, tool wrappers, Trace IR recording, commit validation, recovery, and evaluation logging.

\subsection{Conservative Dependency Tracing}
\label{app:tracing}

Riker wraps Pi's \texttt{bash}, \texttt{read}, \texttt{write}, and \texttt{edit} tools so that execution and tracing share one Linux environment; the sets it yields are what commit, replay, and RAW/write-after-write (WAW) hazard checks consume. Native \texttt{read}, \texttt{write}, and \texttt{edit} expose explicit file dependencies and post-state digests. Tests, compilation, search, and general \texttt{bash} commands are treated conservatively and reuse a result only when the record proves that their dependencies, projected fields, and effects are complete. For these opaque commands, the command runs under \texttt{rkr}, and \system lowers the file events Riker records---\texttt{PathRef}, \texttt{MatchContent}, \texttt{MatchMetadata}, and \texttt{UpdateContent}---into the read set $R_i$, absence set $A_i$, write set $W_i$, and effect set $E_i$ of the Trace IR record, then drops transient accesses to \texttt{/tmp}, process-private paths, and entries the wrapper schema declares ignorable. Each read-set entry carries a concrete sandbox-state digest captured at trace summary---file bytes and mode, directory state, or path-absent---which $V_{\mathrm{dep}}$ recomputes on committed state and compares at the frontier. Coverage is the local process tree Riker traces: ordinary \texttt{fork} and \texttt{exec} children are included, so a shell command and the processes it spawns contribute to one dependency set. It does not extend to detached daemons, the internals of external database services, sandboxed multi-process browsers, remote services, or network state. A call whose dependencies reach outside that tree is a speculation barrier or, when it reads a long-running service, is held on that service's loaded-version record rather than traced into.

Following Riker~\cite{curtsinger2022riker}, path dependencies are recorded conservatively. A path passed to an operation such as \texttt{open}, \texttt{stat}, or \texttt{rename} counts as an access whether or not the operation succeeded. A failed lookup is recorded in the absence set $A_i$, so that a later committed effect creating that path invalidates the candidate. Parent directories are tracked because a directory update can change which file a child path resolves to. Symbolic-link use records both the link path and the target path when either can affect the result. Some inputs fall outside these rules: directory enumeration, environment state, process identity, and time or randomness sources. If one of them can affect a result and cannot be captured, the call becomes a speculation barrier rather than a record with partial evidence.

\subsection{Overlay Promotion and Replay}
\label{app:promotion}

The fast commit path is overlay promotion: when the frontier action matches a candidate and the dependency, result, and effect checks pass, \system promotes that overlay to become the committed sandbox. If the predicted observation also matches the sandbox observation under the wrapper's canonicalization rule ($V_{\mathrm{pred}}$, Section~\ref{sec:squash}), descendants that consumed it remain eligible; otherwise only the current result is reused. If promotion is unavailable but a replayable Trace IR entry exists for the same canonical frontier action, \system clones a replay overlay, applies the stored effect, and checks the post-state digest; replay is sequential and rejects on RAW/WAW hazards. Otherwise the call runs freshly in the committed sandbox as ordinary execution.

\subsection{Trace IR Schema and Validation}
\label{app:trace-ir-schema}

\begin{figure}[t]
\centering
\includegraphics{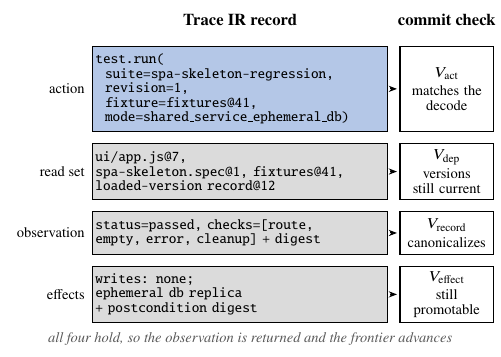}
\caption{Trace IR record from a measured test execution and the commit check each field must pass. Predicates are defined in Section~\ref{sec:commit}.}
\Description{The left column lists four fields of one Trace IR record: the canonical action, which is a test.run call with suite, revision, fixture and mode arguments; the read set, listing three file versions and the loaded-version record of the process contacted; the observation, a passed status with four named checks and a digest; and the effects, an empty write set with an ephemeral database replica and a postcondition digest. An arrow joins each field to the commit check it must pass in the right column: the action to V-act, the read set to V-dep, the observation to V-record, and the effects to V-effect. A line at the bottom states that all four hold, so the observation is returned and the frontier advances.}
\label{fig:trace-ir-example}
\end{figure}

The record for candidate $i$ is
\[
\rho_i =
\left\langle
\canon{\hat{a}_i}, L_i, R_i, W_i, A_i,
\tilde{o}_i, d^{\mathrm{obs}}_i, d^{\mathrm{post}}_i,
E_i, \pi_i
\right\rangle .
\]
Table~\ref{tab:trace-ir-fields} gives the implementation-level mapping from each field to its commit check. The table is included here to make the record format reproducible without interrupting the design-level presentation in Section~\ref{sec:trace-ir}.

\begin{table*}[t]
\centering
\caption{Trace IR fields and commit checks.}
\label{tab:trace-ir-fields}
\small
\begin{tabularx}{\textwidth}{@{}p{0.13\textwidth}p{0.43\textwidth}X@{}}
\toprule
Field & Records & Commit validation \\
\midrule
$\canon{\hat{a}_i}$ & Canonical tool name, JSON arguments, working directory, and environment. & Match the next serial action. \\
$L_i$ & Committed or speculative parent state used to fork the sandbox. & Lineage precondition of $V_{\mathrm{dep}}$: reject a candidate forked from a superseded or squashed parent. \\
$R_i$ & Read set with versions or content digests. & Reject if an older committed effect changes a resource the tool read. \\
$A_i$ & Absence set: failed path or resource lookups that affected execution. & Reject if an older committed effect creates one of those paths or resources. \\
$W_i,E_i$ & Private writes and deferred effect deltas. & Publish only captured deltas; no unrelated committed state may change. \\
$\tilde{o}_i,d^{\mathrm{obs}}_i$ & Sandbox observation and observation digest. & Validate the recorded observation under the wrapper's canonicalization rule; compare it with predicted observations only to keep descendants eligible. \\
$d^{\mathrm{post}}_i,\pi_i$ & Effect-scoped postcondition digest and promote-or-replay rule. & Confirm the captured effect scope before promotion or replay; otherwise run the tool serially. \\
\bottomrule
\end{tabularx}
\end{table*}

\subsection{Process Loads and Holds}
\label{app:process-loads}

A service-restart action (\texttt{restart}) names the process, the stop signal, the workspace to load, and a readiness path. The wrapper stops the shared process, reloads it from the committed workspace, and polls the readiness path; once the process answers, the wrapper reads back the loaded-version record and publishes it as part of the restart's commit. A test declares in its arguments whether it runs against the shared process or in an isolated workspace; only the former binds to the most recent preceding restart. In the representative 263-call audit, all 46 shared-process tests bind to the most recent restart and the 2 isolated tests bind to none; no other tool class carries a binding. A candidate whose binding is unresolved forks its overlay, then waits; the runtime journal records the hold and the release as separate events, and in the traces the release, publication of the loaded-version record, and the start of execution share one timestamp.

\subsection{Scheduling Inputs and Resource Settings}
\label{app:sched-settings}

Table~\ref{tab:sched-inputs} lists the Run-Ahead Controller's inputs. Expected duration is an exponentially weighted running mean over durations already observed for that tool class in the current task, initialized from registry defaults so no profiling pass is required. The reported runs use chain depth $K=6$, a budget of three live candidates, and at most two forks in progress at once; together with the committed sandbox and replay overlays this stays within the eight concurrent sandboxes our E2B account allows. Dispatch-order weights $W$ rank long tests over fast reads: the reported runs use 30 for long tests, 5 for short tests, 3 for checks, and 1 for reads. They are heuristic order-of-magnitude estimates of the latency a hit would hide, not measurements and not a learned policy, and low-weight classes can starve when the budget is full of tests. Selection uses no future trajectory: a candidate enters consideration only after the Action Drafter emits its concrete tool action, and the Run-Ahead Controller then applies class, resource, and producer checks. The registry holds one entry per tool class rather than per command---six in the prototype---and a class with no entry is treated as non-speculatable, which costs an opportunity and never correctness. Drafting runs one Observation Drafter request and one Action Drafter request at a time. A speculatable class whose content actually depends on unresolved intermediate state may still be drafted; if the main agent later emits a different canonical action, the candidate fails $V_{\mathrm{act}}$ and costs one discarded overlay.

\begin{table}[t]
\centering
\caption{Run-ahead scheduling inputs. None is a correctness input.}
\label{tab:sched-inputs}
\scriptsize
\begin{tabularx}{\columnwidth}{@{}l X X@{}}
\toprule
Input & Online source & If wrong \\
\midrule
Drafted tool action & Action Drafter output from an online context built from the committed history, optionally extended by prediction-eligible drafted steps & Action mismatch at the frontier; discard \\
Expected duration & Exponentially weighted running mean over that class within the task & Suboptimal wait-or-discard decision at the frontier only \\
Speculatable & Tool registry safety class for the drafted tool action & Call becomes a speculation barrier \\
Loaded-version binding & Test arguments; readiness probe at restart commit & Loaded-version mismatch; discard and serially re-execute \\
\bottomrule
\end{tabularx}
\end{table}

\subsection{Drafter Training and Runtime Settings}
\label{app:drafter-params}

Model variables select the deployed checkpoints. In the reported runs, the main agent runs Qwen3.8-27B. The Action Drafter starts from Qwen3-8B and uses \texttt{checkpoint-256}, the final LoRA adapter. The Observation Drafter uses Qwen-AgentWorld-35B-A3B without task-specific adaptation.

The action-drafter corpus comes from the SWE-Marathon official trajectory release, restricted to trajectories produced by Claude/Codex harness runs and excluding Qwen/Pi runs. After mapping tool calls to Pi canonical actions and discarding trajectories with more than 30\% unmappable steps, 187 of 212 source trajectories remain. A deterministic trajectory-hash split (seed \texttt{20260902}, validation fraction 0.05) yields 177 training trajectories / 51,160 action samples and 10 held-out trajectories / 4,311 depth-1 canonical-action labels. The training data contains no trajectory produced by the \system runtime.

To reproduce the lookahead conditional at chain depth $k$, each training example masks the most recent $k \in \{1,\ldots,6\}$ observations. Labels use the same $\canon{\cdot}$ action representation as commit validation. LoRA rank is 64, $\alpha$ is 128, dropout is 0.05, and adapters are applied to \texttt{q}, \texttt{k}, \texttt{v}, \texttt{o}, \texttt{up}, \texttt{gate}, and \texttt{down\_proj}. Training uses AdamW, learning rate $10^{-4}$, cosine schedule with warmup ratio 0.03, global effective batch size~8 (per-device~1 $\times$ 4~GPUs $\times$ gradient-accumulation~2), one epoch, max sequence length~32,768, and bf16 precision. Training ran for 42\,h on 4$\times$~NVIDIA H20 GPUs. The reported adapter is the latest completed checkpoint; no validation-based early stopping was used.

Each Observation Drafter request provides the parent context, candidate action, tool schema, observation schema version, and environment state digests. Action drafting uses \texttt{topK=2}. Observation drafting uses \texttt{n}-sampling with \texttt{topK=2} and returns the top-scoring observation as the successor drafter context; it does not gate whether that context may feed another Action Drafter call, since the prediction is only compared against the real observation later, at the frontier. Chain depth is capped at 6 (\texttt{maxDraftDepth}). Token budgets are 8192/4096/2048 for the main agent, Action Drafter, and Observation Drafter, respectively. Drafting and preparation latency, measured from selecting an eligible drafter context to having a concrete candidate ready to execute, is reported in Section~\ref{sec:eval-overheads}. The reported runs are fresh online executions rather than replays of a stored reference trajectory; \system records the canonical actions the main agent actually emits and validates candidates against them at the commit frontier.

\subsection{Additional Evaluation Results}
\label{app:eval-extra}

This appendix expands the Evaluation with detailed SWE-Marathon wall-clock
rows, out-of-order issue diagnostics, drafter sweeps, main-agent parity, the
operand-model ablation behind the two-class split of
Section~\ref{sec:versioning}, and the run-ahead allocation sweep behind the
roofline of Section~\ref{sec:eval-roofline}.

\begin{table}[t]
\centering
\caption{End-to-end wall-clock latency on the SWE-Marathon online subset.
\textit{Cap} marks the E2B session limit. \textit{Eq-prog.} entries are
equal-progress times: the configuration has reached the committed-tool-result
count that Serial Pi had at the 6.0\,h cap, not task completion. \textit{Actions}
counts recorded tool actions. \textit{Tool} is the share of Serial Pi wall clock
spent inside tool executions.}
\label{tab:eval-wallclock}
{\setlength{\tabcolsep}{2pt}%
\scriptsize
\begin{tabularx}{\columnwidth}{@{}X c c c r r@{}}
\toprule
Task & Serial & PASTE & \system{} & Actions & Tool \\
\midrule
BioFabric rust & 6.0h cap & 5.3h eq-prog. & 5.1h eq-prog. & 7{,}677 & 22.4\% \\
Embedding eval & 59m & 44.7m & 39.0m & 1{,}036 & 48.5\% \\
Excel clone & 39m & 36.9m & 36.0m & 4{,}513 & 10.9\% \\
Network align. & 6.0h cap & 4.1h eq-prog. & 3.4h eq-prog. & 3{,}494 & 62.7\% \\
JAX$\rightarrow$PyTorch & 3.2h & 2.5h & 133.8m & 1{,}897 & 43.3\% \\
K8s rust & 6.0h cap & 5.4h eq-prog. & 5.2h eq-prog. & 4{,}100 & 19.7\% \\
Mastodon clone & 23m & 22.0m & 21.6m & 301 & 8.9\% \\
NextJS$\rightarrow$Vite & 94m & 74.0m & 66.0m & 4{,}955 & 42.6\% \\
Parameter golf & 118m & 74.3m & 56.9m & 1{,}315 & 74.0\% \\
Ruby$\rightarrow$Rust & 6.0h cap & 5.7h eq-prog. & 5.6h eq-prog. & 619 & 10.5\% \\
Rust C compiler & 6.0h cap & 4.6h eq-prog. & 4.1h eq-prog. & 567 & 45.4\% \\
Rust Java LSP & 6.0h cap & 5.4h eq-prog. & 5.1h eq-prog. & 631 & 20.6\% \\
S3 clone & 48m & 46.1m & 45.4m & 2{,}390 & 7.7\% \\
Slack clone & 27m & 25.4m & 24.7m & 2{,}134 & 12.0\% \\
Stripe clone & 19m & 18.0m & 17.6m & 1{,}713 & 10.4\% \\
VLIW kernel & 43m & 42.0m & 41.7m & 392 & 4.4\% \\
WASM SIMD & 6.0h cap & 5.6h eq-prog. & 5.5h eq-prog. & 2{,}229 & 11.9\% \\
Zstd decoder & 60m & 58.9m & 58.5m & 685 & 3.6\% \\
\bottomrule
\end{tabularx}
}
\end{table}

\paragraph{Per-task distribution.}
Table~\ref{tab:eval-wallclock} gives per-task wall clock for all 18
SWE-Marathon tasks; read as speedup against Serial Pi, it has the following
shape. Every task is at or above Serial Pi, and the spread runs from
1.03$\times$ to 2.07$\times$. The ranking follows the amount of tool time
available to hide more than task length: the six tasks whose Serial Pi runs
spend at least 40\% of wall clock in tools have a 1.60$\times$ geometric-mean
speedup, while the nine tasks below 15\% have a 1.07$\times$ geometric mean.
The four highest points are Parameter golf at 2.07$\times$, Network alignment
at 1.78$\times$, Embedding eval at 1.51$\times$, and Rust C compiler at
1.47$\times$. The four lowest are Zstd decoder at 1.03$\times$, VLIW kernel at
1.03$\times$, S3 clone at 1.06$\times$, and Mastodon clone at 1.07$\times$.

The seven timeout-capped tasks enter as matched-progress ratios and span
1.08$\times$ to 1.78$\times$, inside the range of the completed tasks rather
than above it. The capped metric neither sets the maximum nor lifts the
distribution, so the reported geometric means do not rest on it.

\paragraph{Target-model generality.}
\label{app:eval-target-model}
Table~\ref{tab:eval-target-model} gives the per-task speedups behind the
target-model comparison of Section~\ref{sec:exp_q1}. The DeepSeek~V4~Pro
configuration replaces only the main agent and the Action Drafter adapter:
the drafter is re-fine-tuned on DeepSeek~V4~Pro trajectories mapped to the
same canonical actions, using the corpus construction, masking procedure, and
LoRA hyperparameters of Appendix~\ref{app:drafter-params}; the Observation
Drafter, tool wrappers, sandbox backend, and scheduler settings are
unchanged. Starred rows are matched-progress ratios at the 6.0\,h E2B cap
under the convention of Table~\ref{tab:eval-wallclock}. Seventeen of eighteen
rows agree within 0.03$\times$, and the two configurations share the same
task ranking and 1.24$\times$ geometric mean.

\begin{table}[t]
\centering
\caption{Per-task \system speedup over Serial Pi under two target models.
Starred rows are matched-progress ratios at the 6.0\,h E2B cap.}
\label{tab:eval-target-model}
\scriptsize
\begin{tabularx}{\columnwidth}{@{}X c c@{}}
\toprule
Task & Qwen3.8-27B & DeepSeek V4 Pro \\
\midrule
BioFabric rust$^{\ast}$ & 1.19$\times$ & 1.16$\times$ \\
Embedding eval & 1.51$\times$ & 1.53$\times$ \\
Excel clone & 1.08$\times$ & 1.08$\times$ \\
Network align.$^{\ast}$ & 1.78$\times$ & 1.81$\times$ \\
JAX$\rightarrow$PyTorch & 1.43$\times$ & 1.42$\times$ \\
K8s rust$^{\ast}$ & 1.16$\times$ & 1.14$\times$ \\
Mastodon clone & 1.07$\times$ & 1.06$\times$ \\
NextJS$\rightarrow$Vite & 1.43$\times$ & 1.41$\times$ \\
Parameter golf & 2.07$\times$ & 2.23$\times$ \\
Ruby$\rightarrow$Rust$^{\ast}$ & 1.08$\times$ & 1.06$\times$ \\
Rust C compiler$^{\ast}$ & 1.47$\times$ & 1.46$\times$ \\
Rust Java LSP$^{\ast}$ & 1.17$\times$ & 1.16$\times$ \\
S3 clone & 1.06$\times$ & 1.05$\times$ \\
Slack clone & 1.09$\times$ & 1.08$\times$ \\
Stripe clone & 1.08$\times$ & 1.07$\times$ \\
VLIW kernel & 1.03$\times$ & 1.03$\times$ \\
WASM SIMD$^{\ast}$ & 1.09$\times$ & 1.08$\times$ \\
Zstd decoder & 1.03$\times$ & 1.02$\times$ \\
\midrule
Geometric mean & 1.24$\times$ & 1.24$\times$ \\
\bottomrule
\end{tabularx}

\end{table}

\begin{figure}[t]
\centering
\includegraphics[width=\columnwidth]{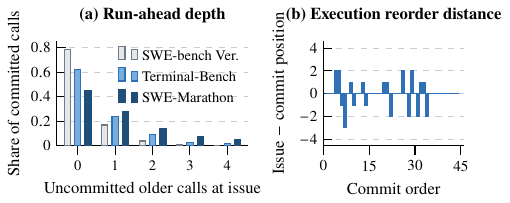}
\caption{Out-of-order issue diagnostics. (a) Run-ahead depth at issue for
committed calls. (b) Execution reorder distance in a traced SWE-Marathon
window.}
\Description{Two panels. Panel a is a grouped bar chart with run-ahead depth
from zero to four on the horizontal axis and share of committed calls on the
vertical axis, with SWE-bench Verified concentrated at depth zero and
SWE-Marathon showing the largest nonzero depths. Panel b is a bar chart over a
commit-ordered traced window, with positive and negative reorder-distance bars.}
\label{fig:eval-ooo}
\end{figure}

\paragraph{Out-of-order issue and in-order commit.}
\label{app:eval-ooo}
\label{sec:eval-ooo}
Figure~\ref{fig:eval-ooo} reports the diagnostic view that supports the
Design Attribution result of Section~\ref{sec:eval-mechanisms}. Panel~(a)
counts how far a committed call had run ahead of the frontier when it issued:
SWE-Marathon has the largest nonzero depth mass, Terminal-Bench~2.0 is
shallower, and SWE-bench Verified stays closest to the serial frontier. These
measurements are intentionally diagnostic rather than a separate result claim:
the paper's claim is not that reordering exists, but that the reordering
survives validation and improves latency relative to same-turn concurrency and
one-step prefetching.

Panel~(b) shows one traced SWE-Marathon window with nonzero execution reorder
distance. Bars above zero are calls that issued before older calls committed;
bars below zero are delayed by operands or resources. The commit frontier still
advances in trajectory order. This is the execution pattern the correctness
argument permits: \system may start work early, but it publishes observations,
effects, and workspace changes only after the corresponding frontier action and
commit predicates validate.

\begin{figure}[t]
\centering
\includegraphics[width=\columnwidth]{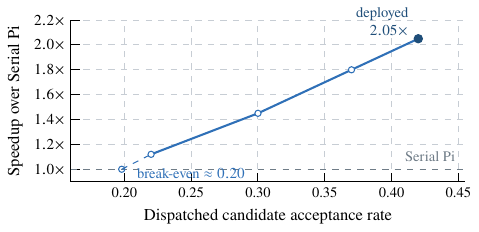}
\caption{Speedup as a function of dispatched-candidate acceptance across
Action Drafter variants.}
\Description{A line chart with dispatched-candidate acceptance rate on the
horizontal axis and speedup over Serial Pi on the vertical axis. Four points
rise from 0.22 acceptance and 1.12 times speedup to 0.42 acceptance and 2.05
times speedup. A dashed fit extrapolates break-even near 0.20 acceptance.}
\label{fig:eval-acceptance}
\end{figure}

\paragraph{Drafter variant sweep.}
\label{app:drafter-variant-sweep}
Table~\ref{tab:eval-drafter-variants} and
Figure~\ref{fig:eval-acceptance} replace the deployed Action Drafter with
weaker checkpoints from the same LoRA run. Acceptance tracks action accuracy
across all four variants and stays below it, as $V_{\mathrm{act}}$ requires.
Speedup is linear in acceptance over the measured range, at 0.47$\times$ per
ten points of acceptance with no sign of saturation, and extrapolating the fit
places break-even near an acceptance rate of 0.20.

\begin{table}[t]
\centering
\caption{Accuracy, acceptance, and latency across Action Drafter variants.
Acceptance counts dispatched candidates accepted at the commit frontier.}
\label{tab:eval-drafter-variants}
\footnotesize
\begin{tabularx}{\columnwidth}{@{}Xccccc@{}}
\toprule
& \multicolumn{2}{c}{Action Drafter} & Obs.\ & Accept- & Speed- \\
\cmidrule(lr){2-3}
Drafter variant & top-1 & top-2 & match & ance & up \\
\midrule
Qwen3.8-8B, no LoRA & 0.29 & 0.44 & 0.52 & 0.22 & 1.12$\times$ \\
Early LoRA checkpoint & 0.36 & 0.53 & 0.58 & 0.30 & 1.45$\times$ \\
Mid LoRA checkpoint & 0.42 & 0.58 & 0.62 & 0.37 & 1.80$\times$ \\
Final LoRA (deployed) & 0.46 & 0.61 & 0.65 & 0.42 & 2.05$\times$ \\
\bottomrule
\end{tabularx}

\end{table}

\paragraph{Main-agent parity.}
\label{app:eval-parity}
Table~\ref{tab:eval-parity} checks the assumption
Section~\ref{sec:design} makes about the main agent. Prompt tokens differ by a
factor of 1.01 between Serial Pi and \system, completion tokens are unchanged,
and total main-agent decode time differs by 1.01. Median time to first token
rises from 310 to 330\,ms and the 95th percentile from 740 to 890\,ms, the cost
of sharing one card with the Action Drafter. The main agent therefore performs
the same decoding work under both configurations, so the wall clock \system
removes comes from tool executions rather than from a faster main agent.

\begin{table}[t]
\centering
\caption{Main-agent parity and speculative resource use.}
\label{tab:eval-parity}
\footnotesize
\begin{tabularx}{\columnwidth}{@{}Xcccc@{}}
\toprule
& Serial Pi & \system & Ratio \\
\midrule
Main-agent prompt tokens (M/task) & \phantom{0}8.2 & \phantom{0}8.3 & 1.01 \\
Main-agent completion tokens (M/task) & \phantom{0}0.41 & \phantom{0}0.41 & 1.00 \\
Main-agent decode time (s/task) & 1{,}640 & 1{,}660 & 1.01 \\
Main-agent TTFT, median (ms) & \phantom{0}310 & \phantom{0}330 & 1.06 \\
Main-agent TTFT, p95 (ms) & \phantom{0}740 & \phantom{0}890 & 1.20 \\
\midrule
Drafter tokens (M/task) & \phantom{0}0 & \phantom{0}1.6 & --- \\
Drafter GPU time (s/task) & \phantom{0}0 & 1{,}480 & --- \\
Sandbox CPU time (s/task) & 2{,}760 & 4{,}310 & 1.56 \\
\bottomrule
\end{tabularx}

\end{table}

\begin{table}[t]
\centering
\caption{Operand ablation. Normalized latency, stale loaded-version rejects and
serial re-executions per task, false accepts, and the share of in-flight
operands held.}
\label{tab:eval-operand}
\footnotesize
\begin{tabularx}{\columnwidth}{@{}X c c c@{}}
\toprule
 & & \multicolumn{2}{c}{Single-version} \\
\cmidrule(lr){3-4}
 & \system{} & Permissive & Conservative \\
\midrule
Normalized latency & 0.58 & 0.68 & 0.85 \\
Stale loaded-version rejects / task & 1.1 & 10.6 & 0 \\
Serial re-executions / task & 2.2 & 12.4 & 2.1 \\
False accepts & 0 & 0 & 0 \\
Held in-flight operands & 15\% & 0\% & 61\% \\
\bottomrule
\end{tabularx}

\end{table}

\paragraph{Operand model ablation.}
Table~\ref{tab:eval-operand} replaces the two-class operand model of
Section~\ref{sec:versioning} with two single-version alternatives, holding the
commit validator, the drafters, the tool wrappers, and the scheduling inputs
fixed. \emph{Single-version permissive} treats every dependency as a ready
operand and never holds a candidate on a producer. \emph{Single-version
conservative} treats every dependency as in flight and holds any candidate
whose inputs a pending write may touch.

Both alternatives are safe and both are slower. Permissive raises normalized
latency from 0.58 to 0.68. With nothing held on a producer, a test drafted
before a service-restart action reads the stale loaded-version record and
$V_{\mathrm{dep}}$ rejects it at the frontier: stale rejects rise from 1.1 to
10.6 per task and serial re-executions from 2.2 to 12.4, so the wait the
permissive model avoids is paid back at commit. Conservative raises normalized
latency to 0.85. Holding 61\% of speculative dependencies rather than 15\%
drives stale rejects to zero, and it also collapses the edit--test window that
Section~\ref{sec:eval-e2e} identifies as the largest source of hideable slack,
leaving 2.1 serial re-executions per task with almost no overlap to show for
them. False accepts are zero in all three columns: safety comes from commit
validation, which none of the variants changes, and the two-class operand model
is what turns that safety into reuse.

\begin{figure}[t]
\centering
\includegraphics[width=\columnwidth]{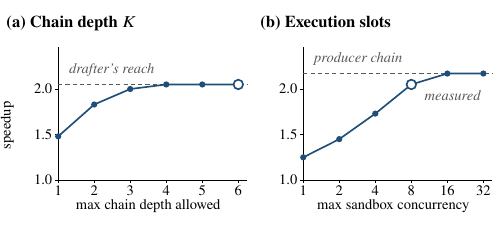}
\caption{Trace-driven sensitivity of speedup to each run-ahead allocation, with
the other allocation held at its reported value. Only the reported
configuration ($K{=}6$, eight slots, 2.05$\times$) is measured; the remaining
points are replay estimates anchored on it.}
\Description{Two small line charts of replay-estimated speedup. Left: speedup against maximum draft depth from one to six, rising from about 1.5 at depth one to about 2.05 at depth four and flat thereafter, with a dashed horizontal line labelled drafter's reach at 2.05 and an open circle at depth six. Right: speedup against maximum sandbox concurrency at 1, 2, 4, 8, 16 and 32 slots, rising from about 1.25 to 2.17 at sixteen slots and flat thereafter, with a dashed line labelled producer chain at 2.17 and an open circle labelled measured at eight slots, the only configuration that was actually run.}
\label{fig:eval-bounds}
\end{figure}

\paragraph{Run-ahead allocation.}
Figure~\ref{fig:eval-bounds} sweeps the two allocations of
Section~\ref{sec:scheduling} one at a time, holding the other at the value the
reported runs use. The sweep is trace-driven rather than a set of fresh runs:
each point replays the recorded dependence graph and per-call durations of the
measured executions under a different allocation, anchored on the one
configuration that was actually run ($K{=}6$, eight slots, 2.05$\times$). We
report it as an estimate of where each allocation stops paying, not as measured
speedup. Under that model, chain depth raises speedup from 1.50$\times$ at
$K{=}1$ to 2.05$\times$ at $K{=}4$ and is flat from there to the $K{=}6$ the
reported runs use. Depth is not the binding allocation: a chain ends at the first failed
predicted observation, and Section~\ref{sec:eval-mechanisms} measures no chain
committing more than four candidates, so raising the cap admits candidates that
the commit frontier discards anyway.

Sandbox concurrency rises from 1.25$\times$ at one slot to 2.05$\times$ at the
eight-slot configuration the reported runs use, which is also the largest
concurrency our E2B account allows (Appendix~\ref{app:sched-settings}); the
sixteen- and thirty-two-slot points lie beyond what the account can run and are
replay estimates only. Under that model speedup reaches 2.17$\times$ at sixteen
slots and is flat beyond, so doubling from the measured eight would add an
estimated 0.12$\times$ and nothing further. Width saturates because the producer
chain still executes in order: once every independent candidate holds a slot, the remaining
serialization is the dependence structure itself, which leaves nothing for the
extra slots to run. The two curves therefore report a separate SWE-Marathon
allocation sensitivity: drafter reach bounds depth, and producer-chain
serialization limits additional sandbox slots. Figure~\ref{fig:eval-limit}
uses the same unlimited-sandbox idea at the three-benchmark aggregate level,
removing the slot cap while retaining dependence serialization.

\end{document}